%% file: root.tex
\documentclass[conference]{IEEEtran}
\usepackage{times}

\usepackage[numbers]{natbib}
\usepackage{multicol}

\usepackage{graphicx} 
\usepackage{wrapfig}

\usepackage[export]{adjustbox}
\usepackage[position=top]{subfig}
\usepackage{subcaption} % Use subcaption for subfigures and subtables

\usepackage{threeparttable}
\usepackage{makecell}
\usepackage{multirow}
\usepackage{booktabs} 

\usepackage{pifont}
\usepackage[ruled, lined, longend, linesnumbered]{algorithm2e}
\usepackage{amsmath}
\usepackage{amssymb}
\usepackage{amsfonts}

\usepackage{xcolor}

\definecolor{violet}{RGB}{148,0,211}
\usepackage[bookmarks=true]{hyperref}
\usepackage{url}

\begin{document}

% \title{Template paper for the \\Robotics: Science and Systems Conference}

\title{UEREBot: Learning Safe Quadrupedal Locomotion under Unstructured Environments and High-Speed Dynamic Obstacles}

% You will get a Paper-ID when submitting a pdf file to the conference system
% \author{Author Names$^{1}$ Omitted for Anonymous Review. Paper-ID \\
% 1 affiliation xxxx, 2 xxx\\
% \href{http:\\www.google.com}{website}
% }

% \author{\authorblockN{Michael Shell}
% \authorblockA{School of Electrical and\\Computer Engineering\\
% Georgia Institute of Technology\\
% Atlanta, Georgia 30332--0250\\
% Email: mshell@ece.gatech.edu}
% \and
% \authorblockN{Homer Simpson}
% \authorblockA{Twentieth Century Fox\\
% Email: homer@thesimpsons.com}
% \and
% \authorblockN{James Kirk\\ and Montgomery Scott}
% \authorblockA{Starfleet Academy\\
% San Francisco, California 96678-2391\\
% Telephone: (800) 555--1212\\

% avoiding spaces at the end of the author lines is not a problem with
% conference papers because we don't use \thanks or \IEEEmembership

\author{
\authorblockN{
Zihao Xu\textsuperscript{1}\quad
Runyu Lei\textsuperscript{1}\quad
Zihao Li\textsuperscript{2}\quad
Boxi Lin\textsuperscript{3}\quad
Ce Hao\textsuperscript{1,4}\quad
Jin Song Dong\textsuperscript{1}
}
\authorblockA{
\textsuperscript{1}National University of Singapore \quad
\textsuperscript{2}ShanghaiTech University \quad
\textsuperscript{3}Xi'an Jiaotong University \\
\textsuperscript{4}Beijing Zhongguancun Academy \\
Homepage: \href{https://uerebot-2026.github.io}{\texttt{https://uerebot-2026.github.io}}
}
}

% \author{Anonymous. Paper-ID [163]}

\makeatletter
\let\@oldmaketitle\@maketitle
    \renewcommand{\@maketitle}{\@oldmaketitle
    \centering
    \includegraphics[width=0.97\textwidth]{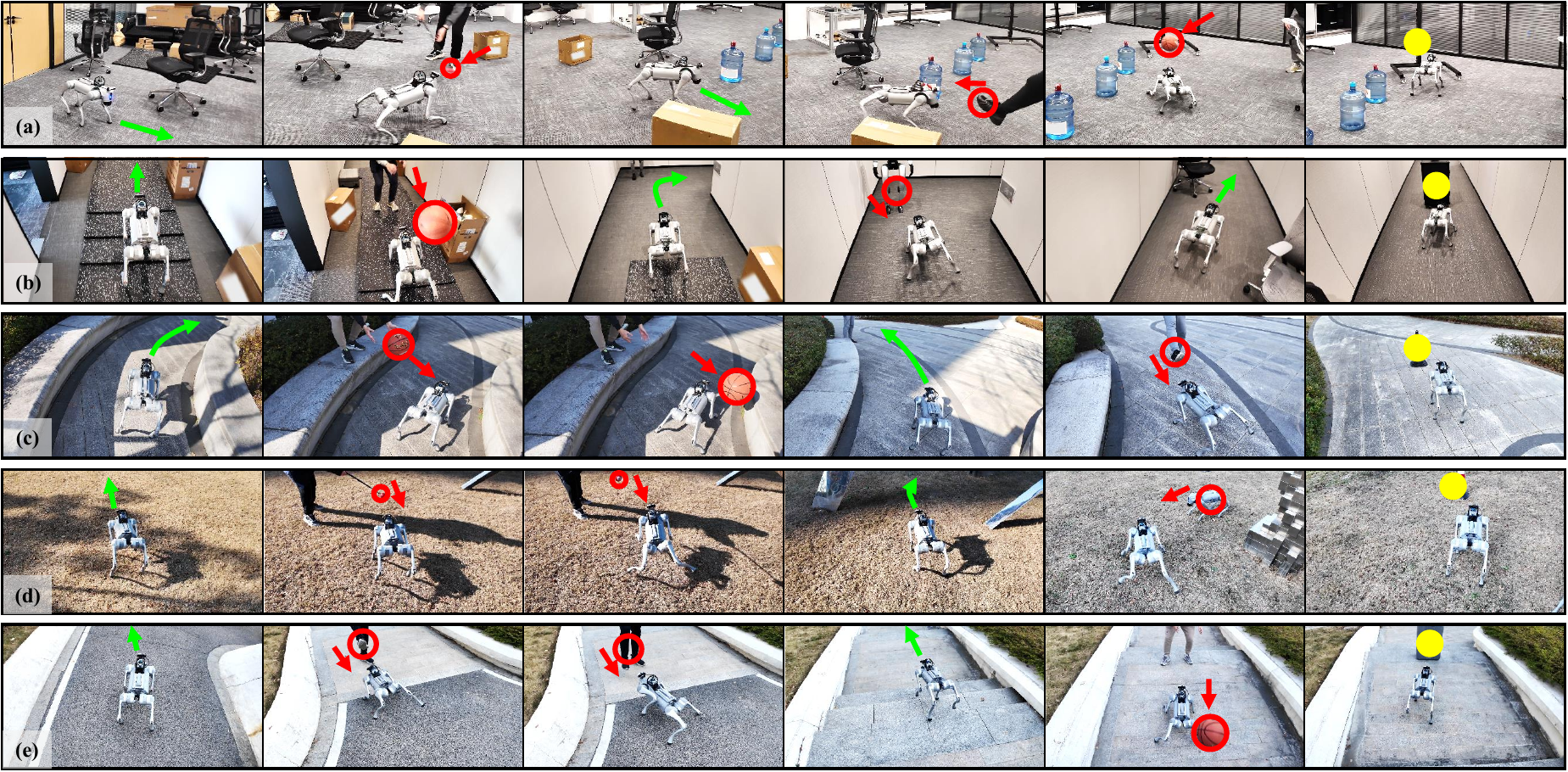}
    % \vspace{-0.5cm}
    \captionof{figure}{
    \textbf{Overview of UEREBot in complex dynamic environments.} The robot tracks a reference path toward the goal region while responding to dynamic attacks under limited reaction time. \textbf{\textcolor{green!80!black}{Green arrows}} indicate the reference direction, \textbf{\textcolor{red}{red arrows}} indicate the attack direction of the dynamic obstacle, and \textbf{\textcolor{yellow!55!orange}{yellow circle}} marks the goal location. (a) Indoor room. (b) Indoor corridor with steps. (c) Outdoor narrow path. (d) Outdoor grass. (e) Outdoor steps. The attacks include poke, hit, and kick, as well as a quadruped and a humanoid robot.}
    % \vspace{-5mm}
    \label{fig:teaser}
    \setcounter{figure}{1}
  % \bigskip
  }
\makeatother

\maketitle

\input{sections/0_abstract}
\input{sections/1_introduction}
\input{sections/2_related_works}

\input{sections/3_prelim}
\input{sections/4_method}

\input{sections/5_experiments}
\input{sections/6_real_robot}

\input{sections/7_conclusion}

\IEEEpeerreviewmaketitle

% \input{sections/10_acknowledgement}

%% Use plainnat to work nicely with natbib. 

\bibliographystyle{plainnat}
\bibliography{references}

\clearpage

\input{sections/11_appendix}

\end{document}

%% file: sections/0_abstract.tex
\begin{abstract}
Quadruped robots are increasingly deployed in unstructured environments. Safe locomotion in these settings requires long-horizon goal progress, passability over uneven terrain and static constraints, and collision avoidance against high-speed dynamic obstacles. A single system cannot fully satisfy all three objectives simultaneously: planning-based decisions can be too slow, while purely reactive decisions can sacrifice goal progress and passability. To resolve this conflict, we propose UEREBot (Unstructured-Environment Reflexive Evasion Robot), a hierarchical framework that separates slow planning from instantaneous reflexive evasion and coordinates them during execution. UEREBot formulates the task as a constrained optimal control problem blueprint. It adopts a spatial--temporal planner that provides reference guidance toward the goal and threat signals. It then uses a threat-aware handoff to fuse navigation and reflex actions into a nominal command, and a control barrier function shield as a final execution safeguard. We evaluate UEREBot in Isaac Lab simulation and deploy it on a Unitree Go2 quadruped equipped with onboard perception. Across diverse environments with complex static structure and high-speed dynamic obstacles, UEREBot achieves higher avoidance success and more stable locomotion while maintaining goal progress than representative baselines, demonstrating improved safety--progress trade-offs.
\end{abstract}

%% file: sections/1_introduction.tex
\section{Introduction}~\label{Sec: Intro}
\vspace{-4mm}

Quadruped robots are increasingly deployed in real-world settings such as industrial plants, warehouses, and outdoor inspection tasks~\cite{liu2024safe,miki2022learning,kim2025high,zhang2024learning}. Their legged morphology enables mobility beyond the capabilities of wheeled platforms~\cite{xiao2025learning,han2025omninet,lu2025fr}. In unstructured environments, safe locomotion involves three tasks: long-horizon goal progress, passability through uneven terrain and static constraints, and collision avoidance against high-speed dynamic obstacles under limited reaction time (Fig.~\ref{fig:teaser}). Prior work typically approaches them with map-based planning~\cite{miki2022learning}, terrain- and constraint-aware navigation~\cite{zhang2025motion,chen2025learning}, or fast reactive avoidance~\cite{xu2025rebot}.

However, goal progress, passability, and high-speed collision avoidance cannot be fully satisfied by a single system under limited reaction time. Goal progress favors planning with long-horizon intent. Passability requires feasibility reasoning over non-convex terrain and static constraints. Collision avoidance against high-speed dynamic obstacles demands instantaneous reactions. The conflict arises because these tasks require different decision speeds within the same real-time control loop. Planning-based decisions can be too slow~\cite{liu2024td3,hewawasam2022past}, while reactive-based decisions can sacrifice goal progress and passability~\cite{xu2025rebot}. This calls for a cohesive spatial--temporal mechanism beyond planning or reactive avoidance.

\begin{figure}[t]
    \centering
    \includegraphics[width=0.97\columnwidth]{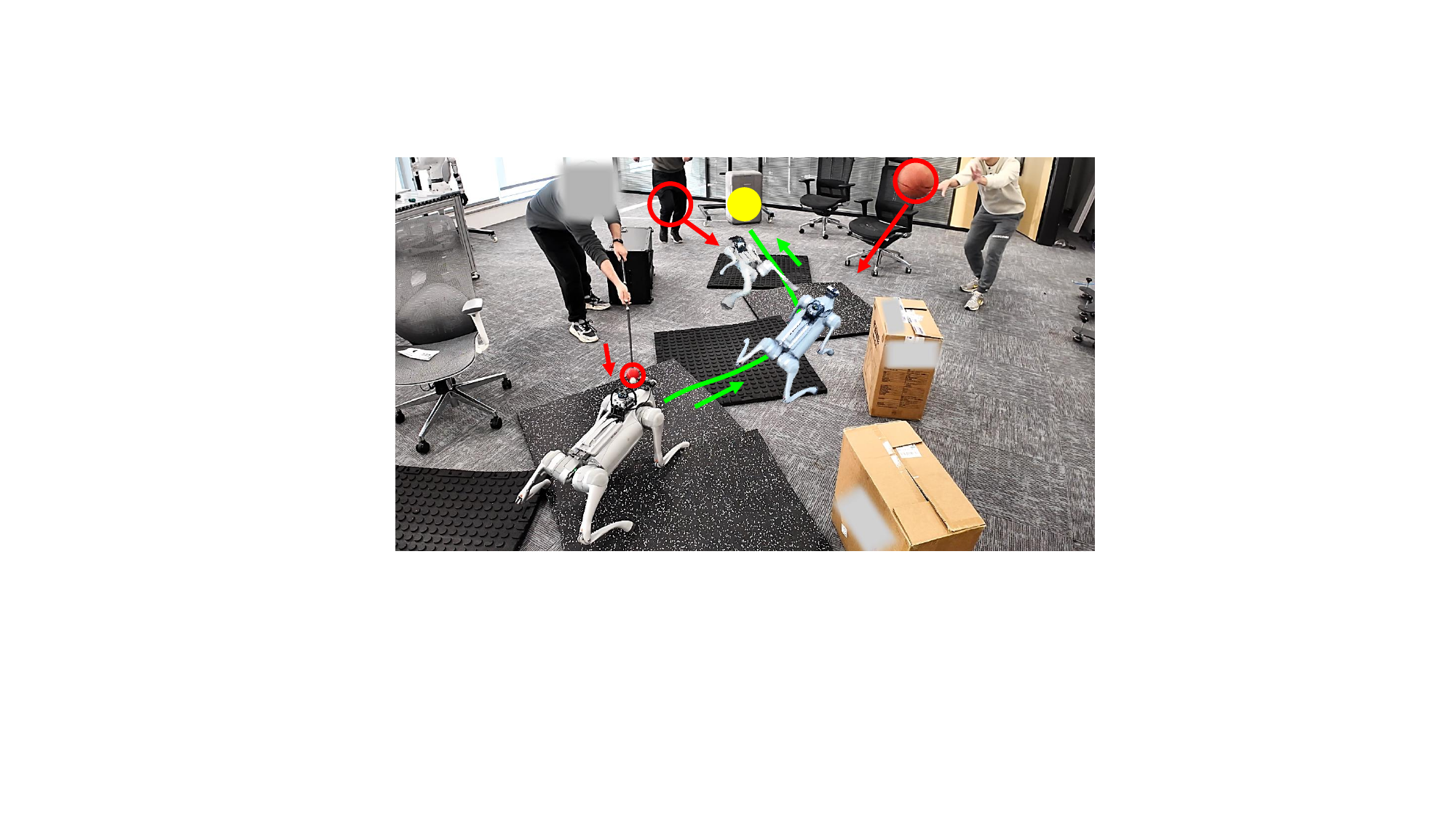}
    \caption{\textbf{Task overview.} The robot tracks a reference path toward the goal region in a complex environment with rough terrain and static obstacles. Along the path, the robot may face dynamic attacks, including poke, hit, and kick.}
    \label{Fig: Preli}
\vspace{-5mm}
\end{figure}

To resolve this conflict, we propose \textbf{UEREBot} (Unstructured-Environment Reflexive Evasion Robot), a hierarchical framework for safe locomotion under unstructured environments and high-speed dynamic obstacles. The key idea is to separate slow planning and instantaneous reflexive evasion, and coordinate them. UEREBot formulates the task as a constrained optimal control problem (OCP) blueprint. It adopts a spatial--temporal planner outputting a reference path to the goal, obstacle predictions, and a threat score. A navigation policy tracks the reference path to maintain goal progress. A reflex policy based on REBot~\cite{xu2025rebot} performs evasion against high-speed obstacles. A threat-aware handoff fuses navigation and reflex actions into a command, and a control barrier function (CBF) shield is a final safeguard.

We evaluate UEREBot to verify that it can maintain goal progress and stable locomotion while avoiding high-speed dynamic obstacles. We conduct experiments in Isaac Lab simulation and deploy UEREBot on a Unitree Go2 equipped with onboard perception~\cite{xu2025aprebot}. Experiments cover complex terrain, static obstacles and high-speed dynamic obstacles. Compared with representative baselines, UEREBot achieves higher avoidance success and larger clearance margins while maintaining stable locomotion and goal progress, demonstrating improved safety--progress trade-offs. Ablation studies further verify the contribution of the key components.

In summary, this paper makes the following contributions:

\begin{itemize}
    \item We formulate the task of safe quadrupedal locomotion under unstructured environments and high-speed dynamic obstacles as a constrained OCP blueprint with coupled objectives and constraints.

    \item We propose UEREBot, a hierarchical framework motivated by the constrained OCP blueprint, which integrates a learned spatial--temporal planner, coordinates navigation and reflex policies via a threat-aware handoff, and adopts a CBF shield as final safeguard.

    \item We validate UEREBot in simulation and on real robots, demonstrating improved safety--progress trade-offs and task completion over representative baselines, with ablation studies supporting the key components.

\end{itemize}

%% file: sections/2_related_works.tex
\section{Related Works}~\label{Sec: Related Works}
\vspace{-4mm}

\textbf{Navigation and planning for goal progress.}
Navigation and planning for goal progress are mature, with reliable deployments in complex scenes~\cite{chen2025survey,dharmadhikari2020motion,xiao2022motion}. Model-based methods~\cite{da2020novel,zhu2021dsvp} maintain environment representations and enable fast replanning via incremental updates and global search, enabling routing in partially observed spaces~\cite{yang2022far}. Learning-augmented approaches~\cite{wijmans2019dd,xu2025navrl,duan2026causalnav,chen2025online} improve unknown-environment navigation by learning compact context representations~\cite{sima2025macronav,he2024alpha,liang2023context} with hierarchical~\cite{seo2022learning}, multi-scale policies to select paths or subgoals~\cite{liang2024hdplanner,tao2024mobile}. For dynamic environments~\cite{song2022dynavins}, recent work REASAN~\cite{yuan2025reasan,cao2025resple} combines online tracking with reactive commands and safety filtering that reshapes nominal actions. Despite these advances, navigation approaches still handle dynamic obstacles mainly through replanning~\cite{oroko2022obstacle}, which can fail under limited reaction time.

\textbf{Passability in unstructured environments.}
Unstructured environments challenge passability over uneven terrain and static obstacles. Model-based approaches build terrain maps and traversability models, and reason with geometric and contact constraints. They combine foothold~\cite{yu2024fast} or contact planning with whole-body optimization~\cite{xu2023whole,lu2023whole} or MPC~\cite{jiang2025nonlinear,corberes2025perceptive} under collision and friction constraints. Learning-based approaches train perceptive locomotion~\cite{miki2022learning} with reinforcement learning (RL) or imitation learning~\cite{zhang2025motion,liu2023vit,peng2020learning,yang2020data}. They often use 3D or context representations~\cite{miki2024learning,chen2024identifying,frey2022locomotion} and hierarchical policies~\cite{lee2024learning} to improve generalization and Sim2Real transfer. These methods target static passability, while dynamic obstacle avoidance under limited reaction time remains less covered.

\begin{figure*}[t]
    \centering
    \includegraphics[width=0.97\linewidth]{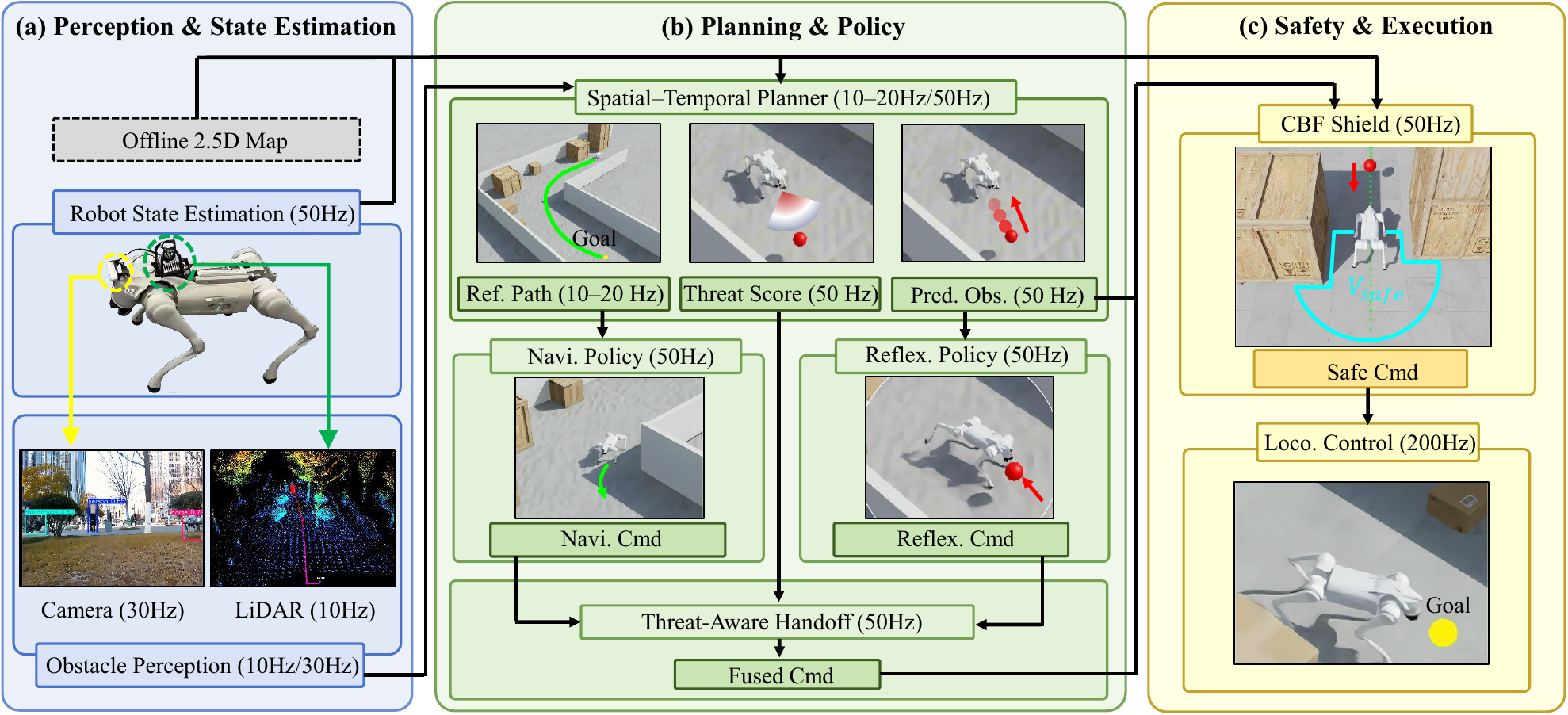}
    \caption{\textbf{Overview of UEREBot framework.} Perception and state estimation provide the robot state and obstacle observations, and an offline 2.5D map is queried to obtain the passability field for terrain and static obstacles. A spatial--temporal planner uses these inputs to produce a reference path, dynamic obstacle predictions, and a threat score. A navigation policy and a reflex policy then generate candidate commands, which are fused by a threat-aware handoff. The fused command is filtered by a CBF shield and executed by the low-level locomotion controller.}
    \label{Fig: pipeline}
\vspace{-5mm}
\end{figure*}

\textbf{Dynamic obstacle avoidance in real time.}
Learning-based obstacle avoidance has been studied for UAVs~\cite{fan2025flying,lu2024fapp,zhang2025threat,lu2022perception}, where single-rigid-body dynamics allow agile policies trained with end-to-end learning for rapid evasion. For legged robots, ABS~\cite{he2024agile} combines locomotion with safety-aware correction to avoid collisions with slower moving obstacles. REBot~\cite{xu2025rebot} learns instantaneous reflexive evasion actions and recovery behaviors to handle high-speed obstacles under limited reaction time. However, REBot primarily focuses on evasion and does not address long-horizon goal progress. Overall, reflexive evasion methods emphasize short-term safety, while integration with long-horizon goal progress remains underexplored.

%% file: sections/3_prelim.tex
\section{Preliminary}~\label{Sec: Prelim}
\vspace{-4mm}

\textbf{Problem formulation.}
We consider quadruped robots that make progress toward a goal region $G$ while maintaining stable locomotion in unstructured environments and avoiding high-speed dynamic obstacles (Fig.~\ref{Fig: Preli}). The robot is described by the state $\tilde{\mathbf{x}}^R=(x,y,\theta,\dot x,\dot y,\dot\theta)$ and a body-frame velocity command $\mathbf{u}=[v_x\ v_y\ \omega]^\top$. We denote the robot position by $\mathbf{p}^R(t)$, expressed in the world frame $\mathcal{W}$ by default unless otherwise specified, and consider a finite horizon $t\in[0,T]$.

We formulate this problem as a constrained OCP blueprint that minimizes a generic locomotion objective $J$ over $[0,T]$:
\begin{equation}
\label{eq:ocp}
\min_{\mathbf{u}(\cdot)} \; J=\int_0^T \ell\big(\tilde{\mathbf{x}}^R(t),\mathbf{u}(t)\big)\,dt,
\end{equation}
where $\ell(\cdot)$ is a locomotion cost that promotes goal progress and stable motion. The OCP is subject to three coupled objectives and constraints that encode goal progress, environment passability, and dynamic-obstacle safety. Goal progress drives $\mathbf{p}^R(t)$ toward the goal region $G$. Environment passability is represented by a signed-distance-like field $\Phi_{\mathrm{env}}(\mathbf{p})>0$, where
$\Phi_{\mathrm{env}}(\mathbf{p})=\min\!\left\{\Phi_{\mathrm{terr}}(\mathbf{p}),\ \Phi_{\mathrm{static}}(\mathbf{p})\right\}$. The robot is feasible under terrain and static constraints if $\Phi_{\mathrm{env}}(\mathbf{p}^R(t))>0$. Dynamic-obstacle safety enforces clearance from high-speed obstacles: for each obstacle $i\in\mathcal{O}$, we require $\Phi_{\mathrm{dyn}}^{(i)}(\mathbf{p},t)>0$, and $\Phi_{\mathrm{dyn}}^{(i)}(\mathbf{p}^R(t),t)>0$ for all $t\in[0,T]$. We use a unified spatial--temporal feasibility field $\Phi_{ST}(\mathbf{p},t)$ to represent these coupled requirements over $[0,T]$.

Due to nonconvex and time-varying spatial--temporal constraints, solving this constrained OCP exactly in real time is intractable under short reaction times; we therefore use it as a design blueprint for a hierarchical framework.

%% file: sections/4_method.tex
\section{Method}~\label{Sec: Method}
\vspace{-4mm}

In this section, we present UEREBot, a hierarchical framework for quadrupedal locomotion under unstructured environments and high-speed dynamic obstacles (Sec.~\ref{Subsec: Framework overview}). The framework consists of a learned spatial--temporal planner (Sec.~\ref{Subsec: Spatial--temporal planner}), a reference-tracking navigation policy (Sec.~\ref{Subsec: Reference-tracking navigation policy}), a reflexive evasion policy (Sec.~\ref{Subsec: Reflexive evasion policy}), a threat-aware handoff policy (Sec.~\ref{Subsec: Threat-aware handoff}), and a CBF safety shield (Sec.~\ref{Subsec: CBF safety shield}).

\subsection{Framework overview}~\label{Subsec: Framework overview}
\vspace{-5mm}

UEREBot is a hierarchical framework for safe locomotion toward the goal region $G$ (Fig.~\ref{Fig: pipeline}). It follows the constrained OCP blueprint to handle uneven terrain and static obstacles, and to avoid collisions with high-speed dynamic obstacles under limited reaction time. The system includes a learned spatial--temporal planner, a navigation policy, a reflex policy, a threat-aware handoff module that fuses the navigation and reflex commands, and a CBF shield that safeguards execution.

At each control cycle, the spatial--temporal planner takes the robot state, the goal region $G$, an offline map encoding terrain and static obstacles, and the current observations of dynamic obstacles. It outputs three items: a reference path toward $G$, predicted states of the dynamic obstacles, and a scalar threat score summarizing the dynamic obstacles. The planner provides intent and threat context, but it does not output control commands. 

Two downstream policies run in parallel. The navigation policy inputs the reference path and robot state, and outputs a velocity tracking command to follow the reference path to $G$ while respecting passability and static constraints. The reflex policy inputs the predicted states of the dynamic obstacles and the robot state, and outputs an evasion command to avoid high-speed dynamic obstacles under limited reaction time.

The threat-aware handoff inputs the robot state, both commands, and the threat score, and outputs a single nominal command by fusing the navigation tracking and the reflexive evasion command. The threat score modulates the fusion, which enables smooth transitions between path tracking and reflexive evasion without discrete switching. The nominal command is then passed to the CBF shield, which enforces safety and actuation constraints. Only the safety-filtered command is sent to the low-level locomotion controller.

\subsection{Spatial--temporal planner}~\label{Subsec: Spatial--temporal planner}
\vspace{-5mm}

The spatial--temporal planner is motivated by the constrained OCP blueprint. The OCP must drive locomotion toward the goal region $G$ while satisfying terrain/static constraints and time-variant dynamic constraints, but it is hard to solve exactly in real time. UEREBot therefore uses a planner to approximate the OCP at the planning level and output the required planning quantities to downstream modules. The planner does not output control commands.

At each cycle $t$, the planner inputs the robot state $\tilde{\mathbf{x}}_t^{R}$, the goal region $G$, the map $\mathcal{M}$ encoding terrain and static obstacles, and the observed dynamic obstacles $\mathcal{S}_t^{O}=\{\mathbf{p}_i^{O}(t),\,\mathbf{v}_i^{O}(t),r_i^{O}\}_{i\in\mathcal{O}}$. It outputs a reference path $\gamma_t^{\mathrm{ref}}$ to $G$, one-step dynamic obstacle predictions $\hat{\mathcal{S}}_{t+1}^{O} = \{\hat{\mathbf{p}}_i^{O}(t+1),\,\hat{\mathbf{v}}_i^{O}(t+1),r_i^{O}\}_{i\in\mathcal{O}}$, and a scalar threat score $\mathcal{T}_t$. The reference path $\gamma_t^{\mathrm{ref}}$ is used by the navigation policy for tracking, the predicted obstacle states $\hat{\mathcal{S}}_{t+1}^{O}$ support reflexive evasion and CBF, and the threat score $\mathcal{T}_t$ is used by handoff.

The planner is implemented as a shared encoder with three learned output heads that generate $\gamma_t^{\mathrm{ref}}$, $\hat{\mathcal{S}}_{t+1}^{O}$, and $\mathcal{T}_t$, and these heads can be trained jointly. We adopt a multi-rate design: $\gamma_t^{\mathrm{ref}}$ is updated at 10--20~Hz, while $\hat{\mathcal{S}}_{t+1}^{O}$ and $\mathcal{T}_t$ are refreshed at 50~Hz to support rapid reflexive evasion, handoff and CBF.

The planner evaluates feasibility using the fields defined in Sec.~\ref{Sec: Prelim}. The map $\mathcal{M}$ is used to construct the passability field $\Phi_{\mathrm{env}}(\mathbf{p})$. For each dynamic obstacle $i\in\mathcal O$, the current obstacle position $\mathbf{p}_i^{O}(t)$ defines a clearance field
\begin{equation}
\Phi_{\mathrm{dyn}}^{(i)}(\mathbf{p},t)=\|\mathbf{p}-\mathbf{p}_i^{O}(t)\|-r^{(i)}_{\mathrm{eff}},
\end{equation}
where $r^{(i)}_{\mathrm{eff}} = r_i^{O} + r^{R} + \delta$ is the safety margin radius. The unified spatial--temporal feasibility field is
\begin{equation}
\Phi_{\mathrm{ST}}(\mathbf{p},t)=\min\Big\{\Phi_{\mathrm{env}}(\mathbf{p}),\min_{i\in\mathcal O}\Phi_{\mathrm{dyn}}^{(i)}(\mathbf{p},t)\Big\}.
\end{equation}
Evaluating at the robot position, $\Phi_{\mathrm{ST}}(\mathbf{p}^R(t),t)>0$ indicates that terrain, static, and dynamic constraints are simultaneously satisfied for the robot at time $t$.

The reference path $\gamma_t^{\mathrm{ref}}$ is obtained by generating a small set of candidate paths toward $G$ on the map $\mathcal{M}$ and selecting the best one with a learned scoring function. Candidates are required to remain feasible under $\Phi_{\mathrm{env}}(\mathbf{p})$ along the path. The selected path is used as the reference for the navigation policy, and we discard a candidate if it would immediately violate dynamic clearance based on $\hat{\mathcal{S}}_{t+1}^{O}$.

\subsection{Reference-tracking navigation policy}~\label{Subsec: Reference-tracking navigation policy}
\vspace{-5mm}

The navigation policy is necessary to realize the goal-progress objective in the constrained OCP blueprint while maintaining stable locomotion. It produces smooth motion toward the goal region $G$ by tracking the reference path $\gamma_t^{\mathrm{ref}}$ from the spatial--temporal planner.

At each control step $t$, the navigation policy $\pi_{\mathrm{nav}}$ inputs the robot state $\tilde{\mathbf{x}}_t^{R}$ and the reference path $\gamma_t^{\mathrm{ref}}$, and outputs a navigation tracking command
\begin{equation}
\mathbf{u}_t^{\mathrm{nav}}=\pi_{\mathrm{nav}}\!\left(\tilde{\mathbf{x}}_t^{R},\,\gamma_t^{\mathrm{ref}}\right).
\end{equation}
The reference path $\gamma_t^{\mathrm{ref}}$ provides a local tracking waypoint and a desired tangential direction, which together define the intended motion direction toward $G$.

We implement $\pi_{\mathrm{nav}}$ as an RL policy. It tracks the local segment of $\gamma_t^{\mathrm{ref}}$ while maintaining well-behaved commands for stable locomotion, including smooth speed regulation and turning behavior. It also learns adaptive locomotion behaviors such as gait and speed adjustment across uneven terrain.

\begin{figure*}[t]
    \centering
    \includegraphics[width=0.97\linewidth]{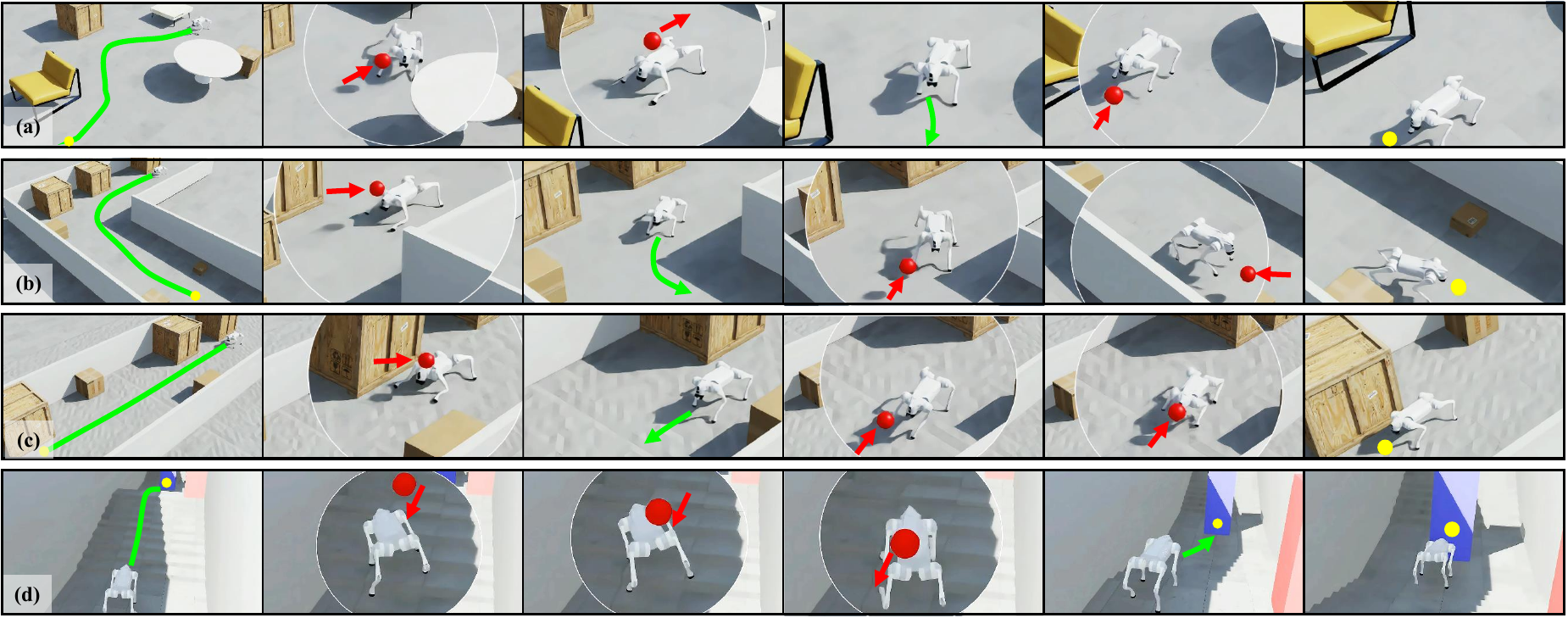}
    \caption{\textbf{Representative simulation experiments.} The robot performs locomotion toward the goal region while traversing rough terrain, passing static obstacles in open and confined spaces, and avoiding high-speed dynamic obstacles under limited reaction time. (a) Open flat ground with static and dynamic obstacles. (b) Narrow corridor. (c) Rough terrain. (d) Stair traversal.}
    \label{Fig: Sim env}
\vspace{-5mm}
\end{figure*}

\subsection{Reflexive evasion policy}~\label{Subsec: Reflexive evasion policy}
\vspace{-5mm}

The reflexive evasion policy is required to satisfy the dynamic-obstacle safety constraint in the constrained OCP blueprint under limited reaction time. It provides low-latency evasive actions to handle high-speed dynamic obstacles that require fast local responses.

At each control step $t$, the reflexive evasion policy $\pi_{\mathrm{refl}}$ inputs the robot state $\tilde{\mathbf{x}}_t^{R}$ and the one-step predicted obstacle states $\hat{\mathcal{S}}_{t+1}^{O}$ produced by the spatial--temporal planner. It identifies the most threatening dynamic obstacle $i_t^\star$ based on the smallest predicted reaction time under the current relative motion, and forms the reflex observation as
\begin{equation}
\mathbf{s}^{\mathrm{refl}}_t = \big[\tilde{\mathbf{x}}_t^{R},\ \hat{\mathcal{S}}_{t+1,i_t^\star}^{O}\big],
\label{eq:reflex_obs}
\end{equation}
where $\hat{\mathcal{S}}_{t+1,i_t^\star}^{O}$ denotes the predicted state of the most threatening dynamic obstacle selected from $\hat{\mathcal{S}}_{t+1}^{O}$.
Given $\mathbf{s}^{\mathrm{refl}}_t$, the reflexive evasion policy outputs a body-frame command
\begin{equation}
\mathbf{u}_t^{\mathrm{refl}} = \pi_{\mathrm{refl}}\!\left(\mathbf{s}^{\mathrm{refl}}_t\right).
\label{eq:reflex_policy}
\end{equation}

The reflexive evasion policy is learned using RL in simulation, with a composite reward function that prioritizes dynamic-obstacle clearance while regularizing motion quality:
\begin{equation}
r_t = r_{\mathrm{safe},t} + r_{\mathrm{reg},t} + r_{\mathrm{ene},t} + r_{\mathrm{rec},t}.
\label{eq:reflex_reward}
\end{equation}
The safety term $r_{\mathrm{safe},t}$ dominates and directly incentivizes clearance from the predicted dynamic obstacle under limited reaction time. The regularization term $r_{\mathrm{reg},t}$ encourages well-behaved commands for stable locomotion during evasive actions. The energy term $r_{\mathrm{ene},t}$ discourages unnecessarily aggressive motions. The recovery term $r_{\mathrm{rec},t}$ encourages rapid recovery to a stable standing posture after evasion, so that the robot can promptly resume stable locomotion.

\subsection{Threat-aware handoff}~\label{Subsec: Threat-aware handoff}
\vspace{-5mm}

The threat-aware handoff reconciles two objectives in the constrained OCP blueprint: tracking the reference path for progress toward $G$ and satisfying the dynamic-obstacle safety constraint under limited reaction time. These objectives are handled by two policies that may produce conflicting commands in the presence of high-speed obstacles. The handoff module coordinates their influence at the command level.

Since the two commands can differ sharply during motion, directly switching from one to the other can introduce discontinuities that destabilize legged locomotion. The handoff instead performs continuous fusion, ensuring smooth transitions between goal tracking and reflexive evasion.

At each control step $t$, the handoff module inputs the navigation command $\mathbf{u}_t^{\mathrm{nav}}$, the reflexive evasion command $\mathbf{u}_t^{\mathrm{refl}}$, and the threat score $\mathcal{T}_t$, and outputs a fused command:
\begin{equation}
\mathbf{u}_t^{\mathrm{fuse}} = f_{\psi}\!\left(\mathbf{u}_t^{\mathrm{nav}},\,\mathbf{u}_t^{\mathrm{refl}},\,\mathcal{T}_t\right),
\end{equation}
where $f_{\psi}$ is a learned fusion function that adjusts the relative influence of the two commands based on $\mathcal{T}_t$. This formulation avoids discrete switching and enables smooth transitions as threats vary.

We train $f_{\psi}$ with RL using rewards that promote fusion:
\begin{equation}
r_t^{\mathrm{handoff}} = r_{\mathrm{coord},t} + r_{\mathrm{smooth},t} + r_{\mathrm{stable},t}.
\label{eq:handoff_reward}
\end{equation}
The coordination term $r_{\mathrm{coord},t}$ encourages consistent fusion, so that the fused command stays close to $\mathbf{u}_t^{\mathrm{nav}}$ when $\mathcal{T}_t$ is low and shifts toward $\mathbf{u}_t^{\mathrm{refl}}$ as $\mathcal{T}_t$ increases. The smoothness term $r_{\mathrm{smooth},t}$ penalizes abrupt changes in $\mathbf{u}_t^{\mathrm{fuse}}$ across control steps. The stability term $r_{\mathrm{stable},t}$ promotes robust locomotion during fusion.

\subsection{CBF safety shield}~\label{Subsec: CBF safety shield}
\vspace{-5mm}

The CBF safety shield serves as the final hard-constraint safeguard in UEREBot, ensuring that the robot’s control commands respect safety constraints. Previous modules are learning-based and optimize soft objectives, so their outputs are not guaranteed to satisfy safety constraints at every step. Accordingly, the CBF shield enforces hard constraints for both static structures and dynamic obstacles, as required by the constrained OCP blueprint.

At each step $t$, the shield inputs the fused command $\mathbf{u}_t^{\mathrm{fuse}}$, the robot state $\tilde{\mathbf{x}}_t^{R}$, the predicted dynamic obstacle states $\hat{\mathcal{S}}_{t+1}^{O}$, and the map $\mathcal{M}$ to query the passability field $\Phi_{\mathrm{env}}(\mathbf{p})$. These inputs are used to define an instantaneous safe command set $V_{\mathrm{safe}}(t)$ that combines static passability constraints, dynamic-obstacle barrier constraints, and actuation bounds. The final executed command is then computed by a safety filter:
\begin{equation}
\mathbf{u}_t^{\mathrm{safe}}=
\begin{cases}
\mathbf{u}_t^{\mathrm{fuse}}, & \mathbf{u}_t^{\mathrm{fuse}}\in V_{\mathrm{safe}}(t),\\
\arg\min_{\mathbf{u}\in V_{\mathrm{safe}}(t)}\|\mathbf{u}-\mathbf{u}_t^{\mathrm{fuse}}\|^2, & \text{otherwise}.
\end{cases}
\end{equation}
The shield does not participate in planning or policy coordination but only intervenes when necessary to prevent safety violations, ensuring the robot's commands remain within the feasible and safe set during execution.

%% file: sections/5_experiments.tex
\begin{table*}[t]
\centering
\caption{Main experiment results.}
\label{tab:sim_main}
\vspace{-0.4em}
\setlength{\tabcolsep}{6pt}
\renewcommand\arraystretch{1.05}
\begin{threeparttable}
\begin{tabular}{lcccccc}
\toprule
Method & GCR $\uparrow$ & ASR $\uparrow$ & TSR $\uparrow$ & PE $\uparrow$ & $d_{\min}$ (m) $\uparrow$ & LC $\downarrow$ \\
\midrule
ABS~\cite{he2024agile} & $0.81 \pm 0.05$ & $0.56 \pm 0.06$ & $0.50 \pm 0.06$ & $0.70 \pm 0.07$ & $0.10 \pm 0.04$ & $0.88 \pm 0.07$ \\
REASAN~\cite{yuan2025reasan} & $\mathbf{0.84 \pm 0.05}$ & $0.61 \pm 0.06$ & $0.54 \pm 0.06$ & $0.74 \pm 0.06$ & $0.12 \pm 0.04$ & $0.75 \pm 0.06$ \\
CWS~\cite{miki2024learning} & $0.83 \pm 0.06$ & $0.47 \pm 0.07$ & $0.41 \pm 0.07$ & $0.77 \pm 0.06$ & $0.06 \pm 0.03$ & $0.82 \pm 0.07$ \\
UEREBot (ours) & $0.80 \pm 0.05$ & $\mathbf{0.81 \pm 0.05}$ & $\mathbf{0.74  \pm 0.05}$ & $\mathbf{0.82 \pm 0.05}$ & $\mathbf{0.17 \pm 0.03}$ & $\mathbf{0.62 \pm 0.05}$ \\
\bottomrule
\end{tabular}
\begin{tablenotes}[flushleft]
\scriptsize
\item Mean $\pm$ std across runs in coupled uneven terrain, static obstacles, and high-speed dynamic obstacles.
\item Higher is better for GCR/ASR/TSR/PE/$d_{\min}$; lower is better for LC. PE/$d_{\min}$/LC are computed over TSR-successful episodes.
\end{tablenotes}
\end{threeparttable}
\vspace{-4mm}
\end{table*}

\section{Simulation Experiments}~\label{Sec: Simulation Experiments}
\vspace{-4mm}

In this section, we evaluate UEREBot in Isaac Lab simulation on tasks that require locomotion toward the goal region $G$ while satisfying terrain/static passability and avoiding high-speed dynamic obstacles under limited reaction time. We present the experiment setup (Sec.~\ref{Subsec: Experiment setup}), followed by the tasks, baselines, and metrics (Sec.~\ref{Subsec: Tasks and baselines}). We then present the main results (Sec.~\ref{Subsec: Main experiments and results}) and analyze the ablations of the threat-aware handoff, the CBF shield, and prediction (Sec.~\ref{Subsec: Ablation studies}).

Our experiments aim to answer the following questions:
\begin{itemize}
    \item \textbf{Q1:} Can UEREBot achieve high task completion and avoidance success under unstructured environments and high-speed dynamic obstacles?
    \item \textbf{Q2:} Can UEREBot achieve improved safety--progress trade-offs compared with representative baselines?
    \item \textbf{Q3:} How do the threat-aware handoff and the CBF shield affect UEREBot performance and execution safety?
    \item \textbf{Q4:} Can UEREBot achieve reliable Sim2Real transfer on a real quadruped platform under realistic sensing noise and latency?
\end{itemize}

\begin{figure}[t]
    \centering
    \includegraphics[width=0.97\columnwidth]{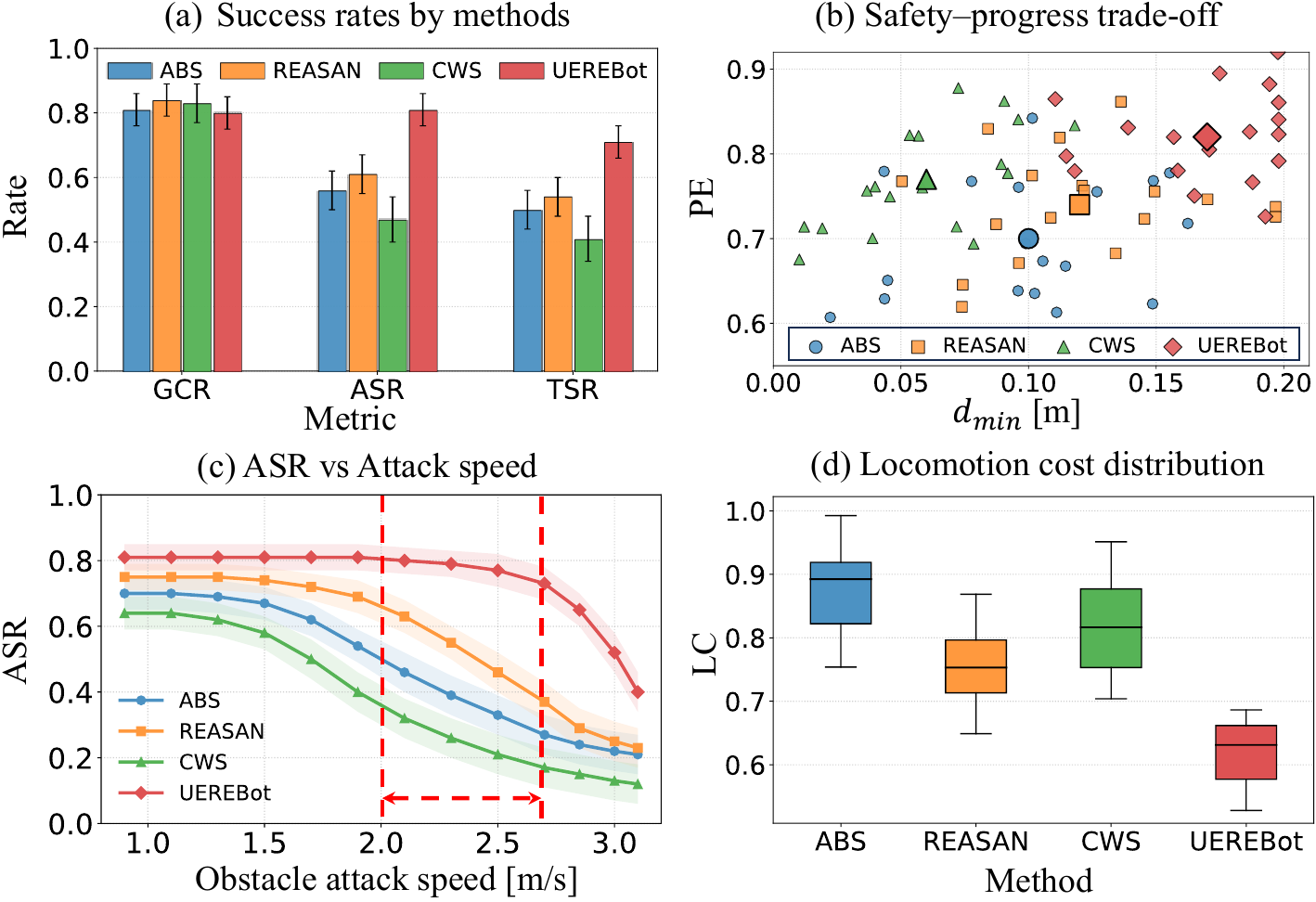}
    \caption{Quantitative results comparing UEREBot against baselines in terms of overall success and its safety–progress trade-off, robustness under increasing attack speed, and the distribution of locomotion cost.}
    \label{Fig: main_results}
\vspace{-5mm}
\end{figure}

\subsection{Experiment setup}~\label{Subsec: Experiment setup}
\vspace{-5mm}

We conduct simulation experiments in Isaac Lab~\cite{mittal2025isaac} using a quadruped robot model that matches the real-robot platform. The robot is commanded by body-frame velocity inputs $\mathbf{u}_t=[v_x\ v_y\ \omega]^\top$, which are tracked by the low-level locomotion controller. We use the resulting base motion and proprioceptive signals as the robot state $\tilde{\mathbf{x}}^R_t$ for the high-level modules.

All experiments use the full UEREBot stack, consisting of a spatial--temporal planner, a reference-tracking navigation policy $\pi_{\mathrm{nav}}$, a reflexive evasion policy $\pi_{\mathrm{refl}}$, a threat-aware handoff module, and a CBF shield. The navigation and reflex policies are pre-trained in simulation and remain fixed during all experiments. The spatial--temporal planner and the handoff module are trained in simulation to generate reference paths and coordinate the actions of the navigation and reflex policies.

We simulate perception by adding noise to dynamic obstacle positions and velocities, and providing obstacle motion predictions. These signals are used by the planner and the CBF shield. Evaluation is carried out on diverse environment layouts, with thousands of evaluation episodes per layout. To ensure a fair comparison, we use the same random seeds across all methods and maintain consistent environmental conditions when comparing with baselines.

\subsection{Tasks, baselines and metrics}~\label{Subsec: Tasks and baselines}
\vspace{-5mm}

\textbf{Tasks.}
We evaluate UEREBot on tasks that require safe locomotion through environments with uneven terrain, static obstacles, and high-speed dynamic obstacles. In each episode, the quadruped must move from a start region to a goal region $G$, while avoiding collisions with obstacles and traversing various terrain. The terrain includes flat ground and more challenging surfaces such as rough terrain, slopes, and steps. Static obstacles, such as furniture and boxes, are randomly placed in the environment. Dynamic obstacles are represented as moving balls, randomized in position, speed, and direction, which the robot must avoid during task execution.

\textbf{Baselines.}
We compare UEREBot with three representative learning-based baselines:
1) \textit{ABS}~\cite{he2024agile} adopts an agile/recovery dual-policy, with a learned reach-avoid value function for risk assessment.
2) \textit{REASAN}~\cite{yuan2025reasan} is a modular framework with separate locomotion, safety-shield, and navigation policies; the safety-shield converts nominal velocity commands into safe commands for reactive avoidance.
3) \textit{CWS}~\cite{miki2024learning} learns locomotion in confined spaces using 3D volumetric scene representations and a hierarchical command-tracking policy.

For fair comparison, all methods use the same robot model and command space, and are evaluated on identical task distributions with the same random seeds.

\begin{table*}[t]
\centering
\caption{Ablation study results.}
\label{tab:abl_ablation}
\vspace{-0.4em}
\setlength{\tabcolsep}{6pt}
\renewcommand\arraystretch{1.05}
\begin{threeparttable}
\begin{tabular}{lcccccc}
\toprule
Variant & GCR $\uparrow$ & ASR $\uparrow$ & TSR $\uparrow$ & PE $\uparrow$ & $d_{\min}$ (m) $\uparrow$ & LC $\downarrow$ \\
\midrule
UEREBot & $\mathbf{0.80 \pm 0.05}$ & $\mathbf{0.81 \pm 0.05}$ & $\mathbf{0.74 \pm 0.05}$ & $\mathbf{0.82 \pm 0.05}$ & $\mathbf{0.17 \pm 0.03}$ & $\mathbf{0.62 \pm 0.05}$ \\
Hard handoff   & $0.74 \pm 0.07$ & $0.71 \pm 0.07$ & $0.63 \pm 0.07$ & $0.76 \pm 0.06$ & $0.13 \pm 0.04$ & $0.73 \pm 0.07$ \\
w/o CBF        & $0.74 \pm 0.07$ & $0.79 \pm 0.06$ & $0.68 \pm 0.07$ & $0.80 \pm 0.06$ & $0.17 \pm 0.04$ & $0.65 \pm 0.06$ \\
w/o pred.      & $0.78 \pm 0.05$ & $0.75 \pm 0.06$ & $0.69 \pm 0.06$ & $0.80 \pm 0.05$ & $0.14 \pm 0.03$ & $0.67 \pm 0.06$ \\
\bottomrule
\end{tabular}
\begin{tablenotes}[flushleft]
\scriptsize
\item Mean $\pm$ std across runs under the same coupled uneven terrain, static obstacles, and high-speed dynamic obstacles as the main experiments.
\end{tablenotes}
\end{threeparttable}
\vspace{-4mm}
\end{table*}

\textbf{Metrics.}
We use system-level metrics to evaluate task completion, safety, and efficiency.
Goal completion rate (\textit{GCR}): $\mathrm{GCR}=N_{\text{goal}}/N_{\text{total}}$, where $N_{\text{goal}}$ counts episodes that reach the goal region $G$ without collision with static obstacles.
Avoidance success rate (\textit{ASR}): $\mathrm{ASR}=N_{\text{avoid}}/N_{\text{total}}$, where $N_{\text{avoid}}$ counts episodes with no collision with any dynamic obstacle.
Overall task success rate (\textit{TSR}): $\mathrm{TSR}=N_{\text{succ}}/N_{\text{total}}$, where $N_{\text{succ}}$ counts episodes that satisfy both the GCR and ASR criteria.

The following metrics are computed over TSR-successful episodes. Path efficiency (\textit{PE}): $\mathrm{PE}=L_{\mathrm{stat}}/L_{\mathrm{act}}$, the ratio of the shortest path on the static map to the actual path length.
Minimum clearance (\textit{$d_{\min}$}): $d_{\min} = \min_{t,\,i\in\mathcal{O}} \Phi_{\mathrm{dyn}}^{(i)}\big(\mathbf{p}^R(t), t\big)$, the minimum clearance margin to any dynamic obstacle.
Locomotion cost (\textit{LC}): the episode-averaged cost that reflects goal progress and stable motion. We compute
\begin{equation}
    \mathrm{LC}=\frac{1}{T}\sum_{t=0}^{T-1}\Bigg(w_g\,\frac{d_G(t)}{d_0} +w_s\,\frac{\|\boldsymbol{\theta}_{rp}(t)\|^2}{\theta_0^2} +w_c\,\frac{\|\Delta\mathbf{u}_t\|^2}{u_0^2} \Bigg),
\end{equation}
where $d_G(t)$ is the distance to the goal, $\boldsymbol{\theta}_{rp}(t)$ denotes the robot's roll and pitch angles, and $d_0$, $\theta_0$, and $u_0$ are reference scales for distance, stability, and control effort, respectively.

\subsection{Main experiments and results}~\label{Subsec: Main experiments and results}
\vspace{-5mm}

Fig.~\ref{Fig: Sim env} shows representative simulation experiments. Across these environments, the robot follows the reference path toward the goal region across uneven terrain such as rough ground and stairs. When a dynamic obstacle approaches, the robot triggers reflexive evasion maneuvers, such as crouching down or jumping away. After evasion, it returns to tracking and completes the task.

\textbf{UEREBot achieves the highest overall success, primarily driven by improved dynamic obstacle avoidance.}
Tab.~\ref{tab:sim_main} and Fig.~\ref{Fig: main_results}(a) show that UEREBot improves ASR by about $20\%$ compared with the best baseline REASAN, and this advantage is correspondingly reflected in TSR with a similar gap. Meanwhile, GCR remains close across methods, indicating that the dominant performance gap is not goal-reaching but achieving collision avoidance against high-speed dynamic obstacles under limited reaction time.

\textbf{UEREBot improves both safety margin and progress efficiency over baselines, yielding a better safety--progress trade-off.}
Tab.~\ref{tab:sim_main} and Fig.~\ref{Fig: main_results}(b) show that UEREBot increases $d_{\min}$ by $0.05\,\mathrm{m}$ over REASAN and improves PE by $0.05$ over CWS, achieving higher safety margin and progress efficiency simultaneously. Moreover, it shows fewer samples at extremely small $d_{\min}$, suggesting that UEREBot maintains larger safety margins in the most safety-critical cases.

\textbf{UEREBot remains more robust as the attack speed increases.}
Fig.~\ref{Fig: main_results}(c) shows that ASR decreases as attack speed increases for all methods, but baselines degrade earlier. In contrast, UEREBot sustains high ASR over a wider speed range and preserves a clear margin in the higher-speed regime, indicating slower performance degradation and lower sensitivity to increased attack speed.

\textbf{UEREBot achieves lower locomotion cost with a tighter distribution.}
Tab.~\ref{tab:sim_main} and Fig.~\ref{Fig: main_results}(d) show that UEREBot reduces LC by $0.13$ compared with baseline REASAN and by up to $0.26$ compared with ABS. The distribution is also more concentrated, indicating less variation across trials. LC is a direct indicator of locomotion effort in our design OCP blueprint, and the notably lower LC shows that UEREBot achieves strong performance while also meeting the cost-related objective emphasized in the OCP blueprint.

\subsection{Ablation studies}~\label{Subsec: Ablation studies}
\vspace{-5mm}

\begin{figure}[t]
    \centering
    \includegraphics[width=0.97\columnwidth]{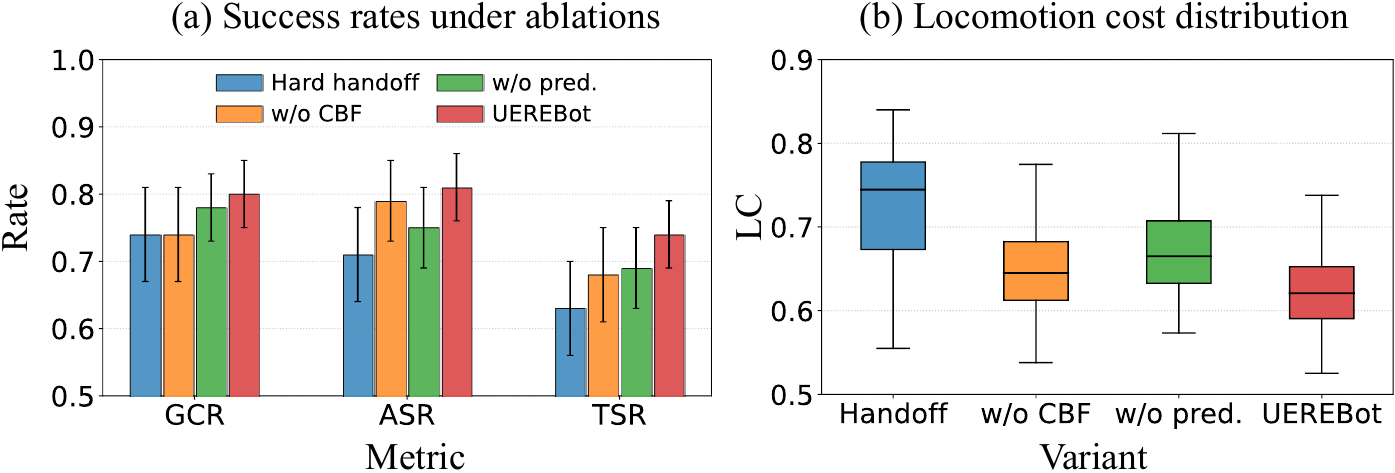}
    \caption{Quantitative results of the ablation study comparing UEREBot against its variants in terms of success rates and the distribution of LC.}
    \label{Fig: ablation_results}
\vspace{-5mm}
\end{figure}

We conduct an ablation study on three key components of UEREBot: the threat-aware handoff that coordinates $\mathbf{u}^{\mathrm{nav}}_t$ and $\mathbf{u}^{\mathrm{refl}}_t$, the CBF safety safeguard, and the prediction module for dynamic obstacles. We construct three variants of the full system: \textit{Hard handoff} replaces the fusion module with a hard switch between $\mathbf{u}^{\mathrm{nav}}_t$ and $\mathbf{u}^{\mathrm{refl}}_t$; \textit{w/o CBF} removes the CBF safeguard; and \textit{w/o pred.} disables prediction by holding the latest obstacle state.

\textbf{Threat-aware handoff ensures smooth navigation--reflex coordination.}
As shown in Tab.~\ref{tab:abl_ablation} and Fig.~\ref{Fig: ablation_results}, hard handoff yields the largest drop in TSR by about $10\%$ and increases LC by roughly $0.1$. This suggests that abrupt switching between $\mathbf{u}^{\mathrm{nav}}_t$ and $\mathbf{u}^{\mathrm{refl}}_t$ introduces command discontinuities that can trigger destabilizing transients in legged locomotion during handoff. As a result, both stable goal progress and reflexive evasion become less reliable, and more corrective actions are required, leading to higher LC.

\textbf{CBF shield prevents reflexive evasion from colliding with static obstacles.}
As shown in Tab.~\ref{tab:abl_ablation} and Fig.~\ref{Fig: ablation_results}, w/o CBF reduces GCR by about $5\%$ and correspondingly lowers TSR by around $6\%$. This drop mainly results from the fact that, under high-speed dynamic obstacle attacks, reflexive evasion without the CBF safeguard can directly collide with nearby static obstacles or enter infeasible terrain. LC also increases slightly, consistent with more frequent corrective actions.

\begin{figure*}[th]
    \centering
    \includegraphics[width=0.97\linewidth]{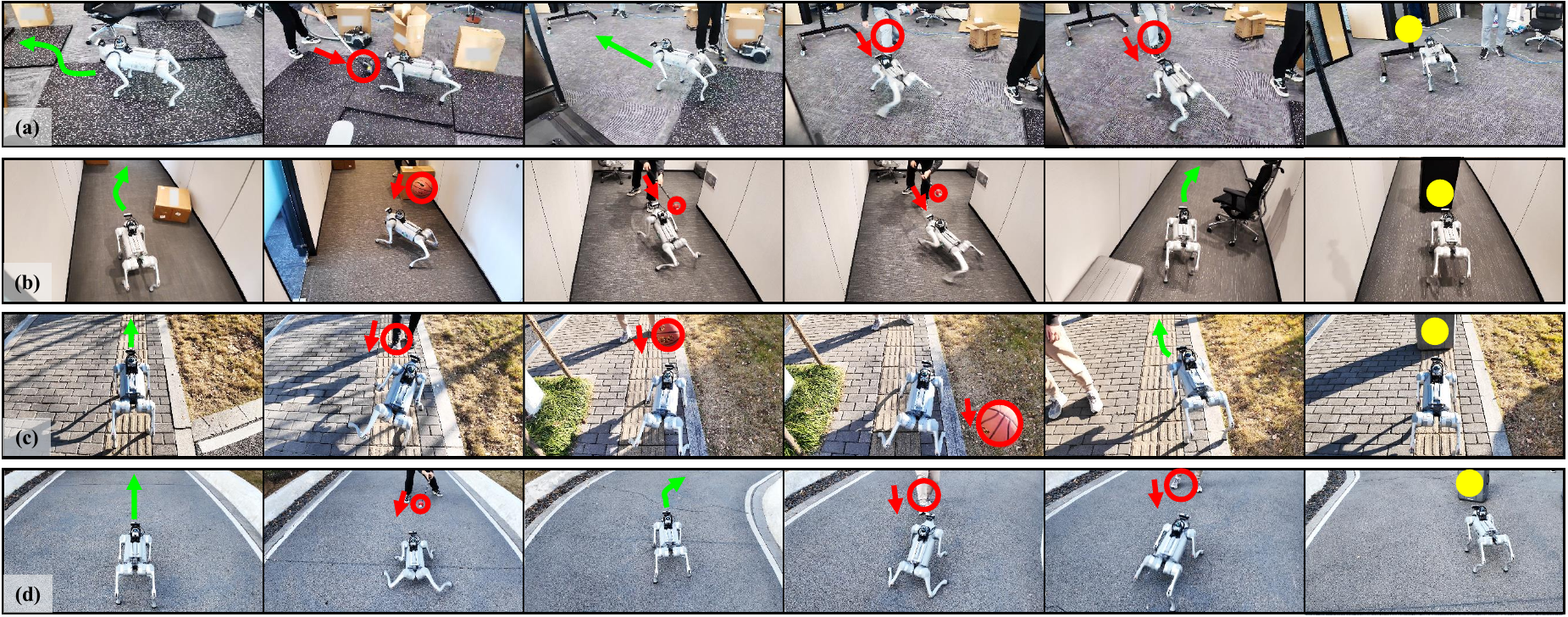}
    \caption{\textbf{Real-robot experiments.} Real-robot trials are conducted in scenarios with varied terrain and layout. (a) Indoor room with rough terrain. (b) Indoor corridor. (c) Outdoor rough terrain. (d) Outdoor slope.}
    \label{Fig: real robot}
\vspace{-5mm}
\end{figure*}

\textbf{Prediction improves reflexive evasion safety margin.}
As shown in Tab.~\ref{tab:abl_ablation} and Fig.~\ref{Fig: ablation_results}(a), w/o pred.\ reduces ASR by about $5\%$ and leads to a clear decrease in $d_{\min}$. This drop is due to delayed reflexive evasion under high-speed attacks when prediction is disabled. The later trigger leaves less reaction time and leads to tighter clearances to the dynamic obstacle.

%% file: sections/6_real_robot.tex
\section{Real-robot Experiments}~\label{Sec: Real-robot}
\vspace{-5mm}

\subsection{Real-robot setup}\label{Subsec: Real-robot setup}

We deploy UEREBot on a Unitree Go2 using a body-frame velocity command interface. We equip the robot with a MID-360 LiDAR and a forward RGB-D camera for onboard perception. The LiDAR provides 360$^\circ$ coverage for omnidirectional perception, while the RGB-D camera covers the frontal field of view. We run the full UEREBot stack onboard in closed loop, with the same modules and interfaces as in simulation.

We conduct 200 real-robot trials in both indoor and outdoor environments. Indoor scenarios include rooms and corridors with static obstacles such as boxes and chairs. Outdoor scenarios include grass, slopes, and steps, together with additional static obstacles in the environment. Dynamic obstacles include thrown balls, fast approaching pedestrians, and close-range stick pokes. In each trial, the robot starts from a predefined region and performs locomotion toward a goal region $G$ while avoiding dynamic obstacles. We evaluate the same metrics as in simulation using the same definitions and setup.

\subsection{Real-robot results and analysis}\label{Subsec: Real-robot results and analysis}

\textbf{Quantitative results.}
We summarize quantitative results over 200 real-robot trials spanning indoor and outdoor settings. UEREBot achieves 75\% GCR, 70\% ASR, and 64\% TSR over all trials. Compared to simulation, these success rates decrease on hardware as expected. LC averaged over TSR-success trials is 0.74, reflecting a higher locomotion cost on hardware than in simulation. Overall, these results confirm that UEREBot transfers reliably from simulation to hardware in the real-robot evaluation.

\textbf{Qualitative analysis.}
Fig.~\ref{Fig: real robot} shows representative real-robot executions in the evaluated scenarios. Across these examples, the robot maintains smooth locomotion toward the goal region and does not get stuck in cluttered areas. When facing fast attacks, it performs reflexive evasion maneuvers, including crouching down, jumping away, and side stepping, while maintaining stable motion. These examples highlight consistent goal progress and reflexive evasion on hardware.

\textbf{Sim2Real gap analysis.}
Compared to simulation, success rates drop and LC increases on hardware. We attribute this Sim2Real gap mainly to three factors. First, the system runs fully onboard with limited computing resources. This increases end-to-end latency and shortens the effective reaction time window under fast, close-range dynamic obstacles. Second, obstacle perception on hardware is affected by noise, missing measurements, and brief occlusions. This can make the estimated dynamic obstacle state less stable and the predictions less reliable. Third, terrain and contact conditions in the real world are more complex than in simulation. This increases tracking errors and corrective actions, which raises LC and can lead to occasional failures. These factors explain the observed performance degradation on hardware relative to simulation.

%% file: sections/7_conclusion.tex
\section{Limitation and Future Works}~\label{Sec: limitation}
\vspace{-5mm}

UEREBot has several limitations. First, our experiments cover a limited set of environments and interaction settings. We do not yet test unknown spaces, denser clutter, or stronger environment changes such as lighting changes, partial occlusion, and low-friction floors. We also do not cover more complex dynamic interactions such as rapid direction changes and repeated attacks from blind spots. Second, real-robot results are reported on one quadruped platform with a fixed sensing and computing setup. Finally, we do not test long-term deployment. In long runs, sensor drift, hardware wear, and changes in the environment may affect performance.

In future work, we will expand the experiments to unknown environments and more complex dynamic interactions. We also plan to validate UEREBot on additional robot platforms with different sensing and computing setups. Meanwhile, we will run longer deployments to assess the impact of sensor drift, hardware wear, and environment changes over time.

\section{Conclusion}~\label{Sec: Conclusion}
\vspace{-4mm}

In this paper, we presented UEREBot, a hierarchical framework for quadrupedal locomotion under unstructured environments with uneven terrain, static constraints, and high-speed dynamic obstacles. UEREBot is motivated by a constrained OCP blueprint and integrates a spatial--temporal planner, reference-tracking navigation policy, reflexive evasion policy, threat-aware handoff, and a CBF shield as a safeguard.

We evaluated UEREBot extensively in simulation and on a real quadruped robot. Across tasks and environments, UEREBot achieves higher avoidance success rate and task success rate than representative baselines. It shows a better safety--progress trade-off, with higher progress efficiency and larger minimum safety margins to dynamic obstacles. Among successful trials, UEREBot achieves lower locomotion cost. Overall, these results indicate that UEREBot provides a practical approach to reliable quadrupedal locomotion in complex dynamic environments.

%% file: sections/11_appendix.tex
\section*{Appendix}~\label{Sec: appendix}
\vspace{-4mm}

\subsection{Notation summary}~\label{Subsec: Notation summary}
\vspace{-4mm}

Tab.~\ref{tab:notation} summarizes the main symbols used in the paper.

\begin{table}[h]
\centering
\caption{Main symbols used in UEREBot.}
\label{tab:notation}
\vspace{-0.4em}
\setlength{\tabcolsep}{7pt}
\renewcommand\arraystretch{1.05}
\begin{threeparttable}
\begin{tabular}{ll}
\toprule
Symbol & Meaning \\
\midrule
$G$ & Goal region \\
$\mathcal{M}$ & Offline 2.5D Map \\
$\mathcal{W}$ & World frame \\
$\tilde{\mathbf{x}}^R$ & Robot state \\
$\mathbf{u}_t$ & Body-frame velocity command \\
$\mathbf{p}^R(t)$ & Robot planar position \\
$\mathcal{S}_t^{O}$ & Observed states of dynamic obstacles in $\mathcal{O}$ \\
$\hat{\mathcal{S}}_{t+1}^{O}$ & Predicted states of dynamic obstacles in $\mathcal{O}$ \\
$\gamma^{\mathrm{ref}}_t$ & Reference path \\
$\mathcal{T}_t$ & Scalar threat score \\
$r^{(i)}_{\mathrm{eff}}$ & Safety margin radius for dynamic obstacle $i$ ($i\in\mathcal{O}$) \\
$\Phi_{\mathrm{terr}}(\mathbf{p})$ & Terrain passability field \\
$\Phi_{\mathrm{static}}(\mathbf{p})$ & Static-obstacle passability field \\
$\Phi_{\mathrm{env}}(\mathbf{p})$ & Environment passability field \\
$\Phi_{\mathrm{dyn}}^{(i)}(\mathbf{p},t)$ & Clearance field for dynamic obstacle $i$ ($i\in\mathcal{O}$) \\
$\Phi_{\mathrm{ST}}(\mathbf{p},t)$ & Spatial--temporal feasibility field \\
$\pi_{\mathrm{nav}}$ & Navigation policy \\
$\mathbf{u}_t^{\mathrm{nav}}$ & Navigation policy command \\
$\pi_{\mathrm{refl}}$ & Reflexive evasion policy \\
$\mathbf{u}_t^{\mathrm{refl}}$ & Reflexive evasion policy command \\
$\mathbf{u}_t^{\mathrm{fuse}}$ & Fused command \\
$V_{\mathrm{safe}}(t)$ & Safe command set \\
$\mathbf{u}_t^{\mathrm{safe}}$ & Executed safe command \\
\bottomrule
\end{tabular}
\end{threeparttable}
% \vspace{-4mm}
\end{table}

\subsection{Details of the problem formulation}~\label{Subsec: Details of the problem formulation}
\vspace{-4mm}

We interpret safe locomotion toward the goal region as minimizing a locomotion cost while remaining feasible under coupled terrain/static constraints and high-speed dynamic obstacles over time. The primary objective is represented by the running cost $\ell\big(\tilde{\mathbf{x}}^R(t),\mathbf{u}(t)\big)$, while feasibility is captured by the spatial--temporal field $\Phi_{\mathrm{ST}}(\mathbf{p}^R(t),t)$ defined in Sec.~\ref{Sec: Prelim}.

An interpretive relaxed Lagrangian view makes explicit how the emphasis can shift from cost efficiency to safety as feasibility margins shrink. We first define the hinge operator
\begin{equation}
[z]_+ \triangleq \max(z, 0).
\label{eq:hinge}
\end{equation}
Using $\Phi_{\mathrm{ST}}(\mathbf{p}^R(t),t)$, the relaxed form is written as
\begin{equation}
\mathcal{L} = \int_0^T \Big( \ell\big(\tilde{\mathbf{x}}^R(t),\mathbf{u}(t)\big) + w(t)\,[-\Phi_{\mathrm{ST}}(\mathbf{p}^R(t),t)]_+ \Big)\, dt .
\label{eq:relaxed_L}
\end{equation}
The violation term $[-\Phi_{\text{ST}}]_+$ activates only when $\Phi_{\text{ST}} \le 0$ and is identically zero whenever $\Phi_{\text{ST}}(\mathbf{p}^R(t),t) > 0$. Accordingly, in feasible regimes \eqref{eq:relaxed_L} reduces to the locomotion cost term, whereas as feasibility margins shrink and violations become imminent, constraint violations contribute an additional penalty.

The time-varying weight $w(t)$ represents changing constraint pressure along execution and is introduced only at the blueprint level to capture the shifting emphasis between cost efficiency and feasibility as conditions become more critical. The relaxed form in \eqref{eq:relaxed_L} is used purely as an interpretive view of the blueprint, clarifying the coupling between locomotion cost and spatial--temporal feasibility that motivates cost-efficient nominal operation together with execution-time safeguarding when necessary.

\subsection{Details of the spatial--temporal planner}~\label{Subsec: Details of the spatial--temporal planner}
\vspace{-4mm}

The spatial--temporal planner is implemented as a multi-head learning module with asynchronous outputs. The reference path $\gamma^{\mathrm{ref}}_t$ is updated at a lower rate ($10$--$20$\,Hz), while the dynamic obstacle prediction $\hat{\mathcal{S}}^{O}_{t+1}$ and the threat score $\mathcal{T}_t$ are updated at a higher rate ($50$\,Hz). Downstream modules always consume the most recent outputs, with $\gamma^{\mathrm{ref}}_t$ held constant between consecutive low-rate updates.

The reference path $\gamma^{\mathrm{ref}}_t$ represents a locally feasible geometric intent toward the goal region and is selected from a fixed set of candidate paths. The planner uses a local 2.5D environment representation constructed offline prior to deployment, from which the environmental feasibility field $\Phi_{\mathrm{env}}$ is derived. Candidate paths are instantiated as motion primitives conditioned on the current robot state. Candidates that violate $\Phi_{\mathrm{env}}(\cdot) > 0$ at any sampled location along the path are discarded. Candidates that would immediately violate dynamic clearance under the current obstacle prediction are also removed. Among the remaining candidates, a learned scoring head ranks the candidates and selects the highest-scoring one as $\gamma^{\mathrm{ref}}_t$. The scoring head is trained with supervision constructed from simulation rollouts, where candidate quality is evaluated based on goal progress, environmental feasibility margins, and motion efficiency. Learning is used here to capture preference among feasible candidates rather than to enforce safety constraints.

The prediction head outputs $\hat{\mathcal{S}}^{O}_{t+1}$, a one-step prediction of dynamic obstacle state aligned with the fast update interval. This prediction is used to compensate for limited perception update rates by providing a time-aligned obstacle state for execution-time modules. The prediction head is trained in simulation using rollout data, where ground-truth next-step obstacle states are available. On the real system, the trained model is used directly to generate time-aligned predictions between perception updates. The prediction output is consumed by the reflexive policy and the CBF safeguard, and does not affect reference path generation.

The threat head outputs a scalar threat score $\mathcal{T}_t$ that summarizes the urgency of dynamic interactions under limited reaction time. The threat score is computed from the shared planner representation and takes the predicted one-step obstacle state as input. Supervision for the threat head is derived from a pressure signal that combines dynamic feasibility margin, relative motion between the robot and obstacles, and the tendency of the interaction to deteriorate within the next control step. As a result, the threat score increases not only when clearance becomes small, but also when obstacles are approaching rapidly. The threat score is updated at the same fast rate as the prediction and is used solely for continuous modulation in the threat-aware handoff; it does not directly determine control actions or override feasibility checks.

\subsection{Details of navigation and reflex policies}~\label{Subsec: Details of navigation and reflex policies}
\vspace{-4mm}

The navigation policy $\pi_{\mathrm{nav}}$ is trained independently to output nominal commands that track the reference path $\gamma^{\mathrm{ref}}_t$ while remaining terrain-executable and cost-efficient. Its per-step reward is defined as
\begin{equation}
r_t^{\mathrm{nav}}=r_{\mathrm{prog},t}+r_{\mathrm{track},t}+r^{\mathrm{nav}}_{\mathrm{stable},t}+r_{\mathrm{effort},t}.
\label{eq:r_nav}
\end{equation}

\noindent\textbf{Path progress.}
Let $\Pi(\mathbf{p})$ denote the projection of planar position $\mathbf{p}$ onto the reference path $\gamma^{\mathrm{ref}}_t$, and let $s(\mathbf{p})$ be the arc-length coordinate of the projected point along the path. We reward incremental advancement along $\gamma^{\mathrm{ref}}_t$ toward the goal region $G$:
\begin{equation}
r_{\mathrm{prog},t}=s\!\left(\mathbf{p}^R(t)\right)-s\!\left(\mathbf{p}^R(t-\Delta t)\right).
\end{equation}

\noindent\textbf{Reference tracking.}
We discourage lateral deviation from the path and encourage heading alignment with the local path direction using exponential shaping of lateral and heading errors:
\begin{equation}
r_{\mathrm{track},t}=\exp\!\Big(-\alpha_p\big\|\mathbf{p}^R(t)-\Pi(\mathbf{p}^R(t))\big\|^2-\alpha_{\psi}\big(\Delta\psi_t\big)^2\Big),
\end{equation}
where $\Delta\psi_t$ is the yaw difference between the robot heading and the path tangent at $\Pi(\mathbf{p}^R(t))$. This term enforces geometric consistency with $\gamma^{\mathrm{ref}}_t$ so that forward progress is achieved near the reference path rather than through large lateral drift.

\noindent\textbf{Terrain-executable stability.}
We promote terrain-executable body motion on uneven terrain by shaping roll--pitch deviation and base angular motion:
\begin{equation}
r^{\mathrm{nav}}_{\mathrm{stable},t}=\exp\!\Big(-\alpha_{\mathrm{rp}}\|\boldsymbol{\theta}_{\mathrm{rp}}(t)\|^2-\alpha_{\omega}\|\boldsymbol{\omega}^R(t)\|^2\Big).
\end{equation}
This term discourages excessive body tilt and aggressive rotations, which are indicative of poorly conditioned locomotion on uneven terrain. By constraining these body-level quantities, the nominal navigation behavior remains terrain-executable.

\noindent\textbf{Cost efficiency.}
We discourage unnecessarily high actuation during nominal navigation by shaping instantaneous mechanical effort:
\begin{equation}
r_{\mathrm{effort},t}=\exp\!\Big(-\alpha_e\sum_j\big|\tau_j(t)\dot q_j(t)\big|\Big).
\end{equation}
This term penalizes strategies that rely on large joint power to obtain progress or tracking, and encourages cost-efficient nominal motion with less aggressive actuation.

The reflexive evasion policy $\pi_{\mathrm{refl}}$ is trained to generate rapid, short-horizon commands that create sufficient separation from high-speed dynamic obstacles under limited reaction time, while remaining physically executable and enabling subsequent recovery. Its per-step reward follows the decomposition in~\eqref{eq:reflex_reward}, and the individual terms are specified as follows.

\noindent\textbf{Safety.}
The primary objective of reflexive evasion is to increase separation from the most threatening dynamic obstacle. We therefore shape the safety reward using the dynamic clearance field $\Phi_{\mathrm{dyn}}^{(i_t^\star)}(\mathbf{p}^R(t),t)$ defined in the main text:
\begin{equation}
r_{\mathrm{safe},t}=\tanh\!\big(\alpha_{\Phi}\,\Phi_{\mathrm{dyn}}^{(i_t^\star)}(\mathbf{p}^R(t),t)\big)-\alpha_v\big(\max(0,v_{\mathrm{app},t})\big)^2.
\end{equation}
The first term provides a continuous incentive to increase clearance, with mild saturation to avoid unbounded scaling. The second term penalizes positive approaching speed $v_{\mathrm{app},t}$, which denotes the radial component of the relative velocity between the robot and the selected threatening obstacle, discouraging evasive actions that continue to reduce separation.

\noindent\textbf{Regularization.}
To ensure that aggressive evasive maneuvers remain physically executable, we include a body-level regularization term:
\begin{equation}
r_{\mathrm{reg},t}=\frac{1}{1+\alpha_{\mathrm{rp}}^{\mathrm{reg}}\|\boldsymbol{\theta}_{\mathrm{rp}}(t)\|^2+\alpha_{\omega}^{\mathrm{reg}}\|\boldsymbol{\omega}^R(t)\|^2}.
\end{equation}
This term favors small roll--pitch deviation and moderate base rotational motion, providing a lightweight executability prior during reflexive evasion.

\noindent\textbf{Energy.}
To avoid achieving evasive separation through unnecessarily high actuation, we penalize instantaneous mechanical effort:
\begin{equation}
r_{\mathrm{ene},t}=\exp\!\Big(-\alpha_{ene}\sum_j\big|\tau_j(t)\dot q_j(t)\big|\Big).
\end{equation}
This term promotes energy-efficient reflexive actions and suppresses impulsive control strategies.

\noindent\textbf{Recovery.}
Finally, we encourage the robot to return to a stable, upright configuration after aggressive evasive maneuvers using a system-level recovery term:
\begin{equation}
r_{\mathrm{rec},t}=\exp\!\Big(-\alpha_h\big(\mathbf{p}^R_z(t)-h^{\mathrm{nom}}\big)^2-\alpha_{\mathrm{rp}}^{\mathrm{rec}}\|\boldsymbol{\theta}_{\mathrm{rp}}(t)\|^2\Big).
\end{equation}
This term promotes restoration of nominal base height and upright posture, facilitating a smooth transition from reflexive evasion back to normal locomotion.

\subsection{Details of threat-aware handoff}~\label{Subsec: Details of threat-aware handoff}
\vspace{-4mm}

The threat-aware handoff is trained with the reward in \eqref{eq:handoff_reward}. In training, we optimize only the handoff module while treating the pretrained navigation and reflex policies as fixed controllers.

\noindent\textbf{Coordination.}
We encourage the fused command to stay close to the navigation output under low threat and to the reflexive output under high threat.
Let $g(\mathcal{T}_t)\in[0,1]$ be a monotone mapping of the threat score. We use exponential shaping of the control discrepancy:
\begin{equation}
\begin{aligned}
r_{\text{coord},t}
&= \bigl(1-g(\mathcal{T}_t)\bigr)\exp\!\Bigl(-\alpha_{\text{nav}}\bigl\|\mathbf{u}^{\text{fuse}}_t-\mathbf{u}^{\text{nav}}_t\bigr\|^2\Bigr) \\
&\quad + g(\mathcal{T}_t)\exp\!\Bigl(-\alpha_{\text{refl}}\bigl\|\mathbf{u}^{\text{fuse}}_t-\mathbf{u}^{\text{refl}}_t\bigr\|^2\Bigr).
\end{aligned}
\label{eq:handoff_coord}
\end{equation}
This term provides continuous, threat-aware guidance for fusion by smoothly biasing $\mathbf{u}^{\text{fuse}}_t$ toward the appropriate source as $\mathcal{T}_t$ increases.

\noindent\textbf{Smoothness.}
We stabilize blending by discouraging abrupt changes in the fused command over time.
Define the control increment $\Delta \mathbf{u}^{\text{fuse}}_t = \mathbf{u}^{\text{fuse}}_t - \mathbf{u}^{\text{fuse}}_{t-\Delta t}$. We apply a bounded shaping term:
\begin{equation}
r_{\text{smooth},t}
= \tanh\!\Bigl(\alpha_{\text{smooth}}\big/\bigl(\|\Delta \mathbf{u}^{\text{fuse}}_t\|+\varepsilon\bigr)\Bigr),
\label{eq:handoff_smooth}
\end{equation}
where $\varepsilon>0$ is a small constant for numerical stability.
This term assigns higher reward to smaller command increments and saturates smoothly to avoid excessive scaling.

\noindent\textbf{Stability.}
We regularize posture during fusion to maintain well-conditioned body configurations.
Using the base roll--pitch vector $\boldsymbol{\theta}_{rp}(t)$, we define
\begin{equation}
r_{\text{stable},t}
= \exp\!\Bigl(-\alpha_{rp}^{\mathrm{stab}}\,\|\boldsymbol{\theta}_{rp}(t)\|^2\Bigr).
\label{eq:handoff_stable}
\end{equation}
This term promotes stable posture throughout fusion, which supports reliable execution of the blended commands.

\begin{figure}[t]
    \centering
    \includegraphics[width=0.97\columnwidth]{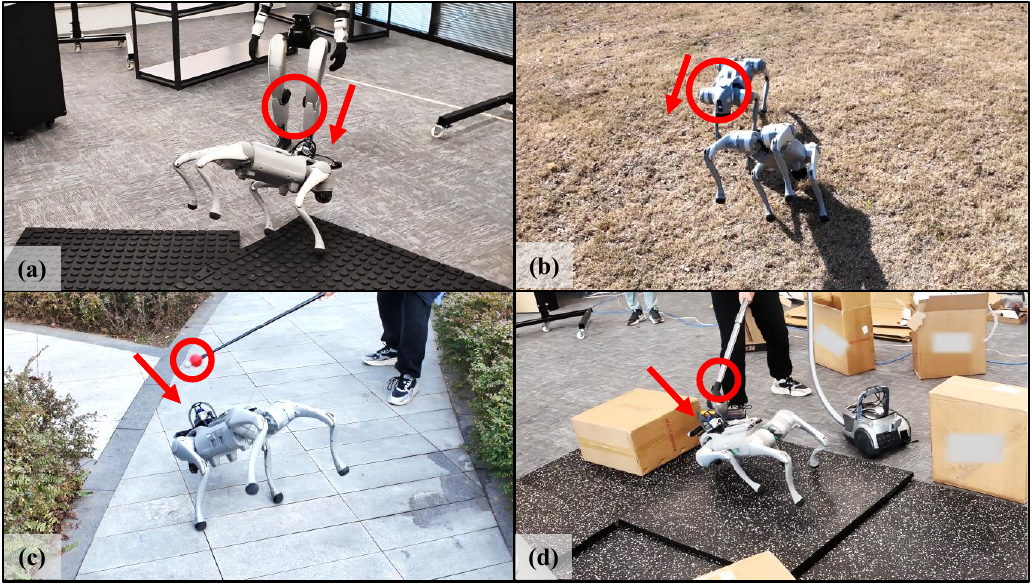}
    \caption{Failure examples with hard handoff. Each subfigure (a--d) shows a representative trial where the hard handoff between navigation and reflex commands causes an unstable transition, leading to loss of balance and a fall.}
    \label{Fig: App_failure_handoff}
\vspace{-5mm}
\end{figure}

We find that hard handoff often causes unstable transitions and may lead to falls (Fig.~\ref{Fig: App_failure_handoff}).

\subsection{Details of the CBF safeguard}~\label{Subsec: Details of the CBF safeguard}
\vspace{-4mm}

\begin{figure}[t]
    \centering
    \includegraphics[width=0.97\columnwidth]{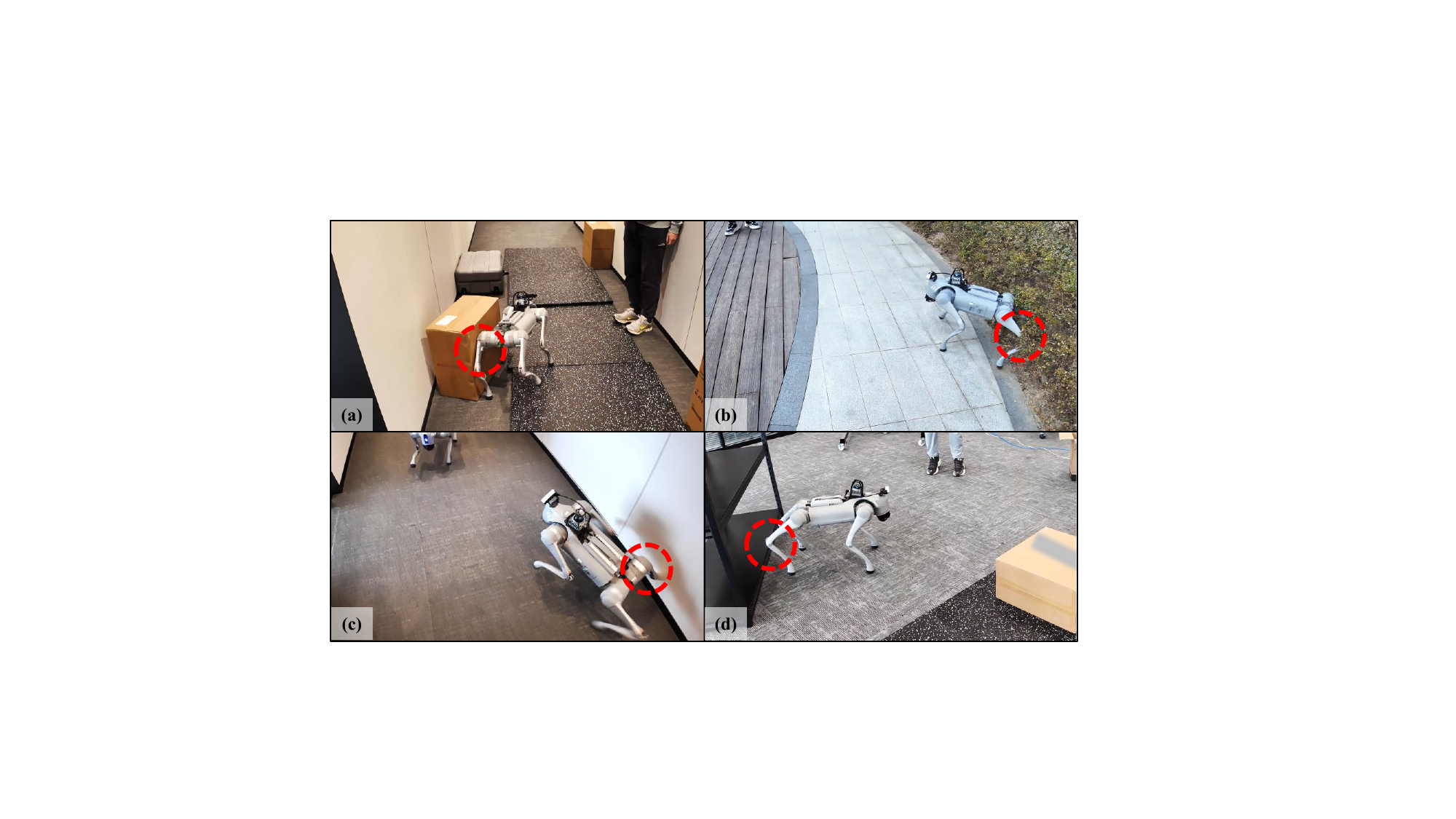}
    \caption{Failure examples without the CBF safeguard. During evasive motion, the robot collides with a nearby static obstacle, highlighting the role of the safeguard in preventing static collisions.}
    \label{Fig: App_failure_CBF}
\vspace{-5mm}
\end{figure}

The main text defines the execution-time safe set $V_{\mathrm{safe}}(t)$ used by the safeguard. Here we summarize a minimal instantiation of $V_{\mathrm{safe}}(t)$ and its CBF constraints. In our experiments, the safeguard is primarily used to prevent unintended contacts with nearby static obstacles during reflexive evasion, while retaining the nominal behavior whenever it remains feasible. Without this safeguard, the robot may collide with nearby static obstacles during evasion (Fig.~\ref{Fig: App_failure_CBF}).

\noindent\textbf{Safe set instantiation.}
We instantiate $V_{\mathrm{safe}}(t)$ as the set of admissible controls that satisfy static feasibility, selected dynamic feasibility, and actuation bounds.
This formulation separates execution safety from the learned coordination in the handoff module, and provides a compact interface for enforcing hard constraints at deployment:
\begin{equation}
V_{\mathrm{safe}}(t)
=\Bigl\{\mathbf{u} \,\Big|\, h_{\mathrm{env}}(t;\mathbf{u})> 0,\ h_{\mathrm{dyn}}^{(i^\star)}(t;\mathbf{u})> 0,\ \mathbf{u}\in\mathcal U\Bigr\},
\label{eq:vsafe_inst}
\end{equation}
where $i^\star$ denotes the selected threatening obstacle index and $\mathcal U$ denotes the actuation bounds.

\noindent\textbf{Barrier functions.}
The static feasibility barrier is constructed from the environment passability field:
\begin{equation}
h_{\mathrm{env}}(t)=\Phi_{\mathrm{env}}\!\bigl(\mathbf{p}^R(t)\bigr),
\label{eq:cbf_h_env}
\end{equation}
which becomes tight when the robot approaches the feasibility boundary induced by the static map.
The dynamic feasibility barrier is instantiated using $\Phi_{\mathrm{dyn}}^{(i)}$ evaluated at time $t$, where $\Phi_{\mathrm{dyn}}^{(i)}$ is computed using the one-step predicted obstacle state provided by the planner:
\begin{equation}
h_{\mathrm{dyn}}^{(i)}(t)=\Phi_{\mathrm{dyn}}^{(i)}\!\bigl(\mathbf{p}^R(t),t\bigr).
\label{eq:cbf_h_dyn}
\end{equation}
In practice, enforcing $h_{\mathrm{dyn}}^{(i)}$ for the selected index $i^\star$ provides a lightweight safeguard against imminent dynamic collisions while keeping the constraint set compact.

\noindent\textbf{CBF constraints.}
For each active barrier, we treat $h(t)$ as a shorthand of $h\!\bigl(\tilde{\mathbf{x}}^R(t),t\bigr)$ and enforce the standard CBF condition
\begin{equation}
\dot{h}(t)+\alpha\,h(t)\ge 0,
\label{eq:cbf_condition}
\end{equation}
with $\alpha>0$.
Under a control-affine approximation of the state dynamics,
$\dot{\tilde{\mathbf{x}}}^R=f(\tilde{\mathbf{x}}^R,t)+G(\tilde{\mathbf{x}}^R,t)\,\mathbf{u}$, the condition in \eqref{eq:cbf_condition} yields an affine inequality constraint on $\mathbf{u}$, which is used to define the admissible set in \eqref{eq:vsafe_inst}.

\subsection{Onboard perception system}~\label{Subsec: Onboard perception system}
\vspace{-4mm}

\begin{figure}[t]
    \centering
    \includegraphics[width=0.97\columnwidth]{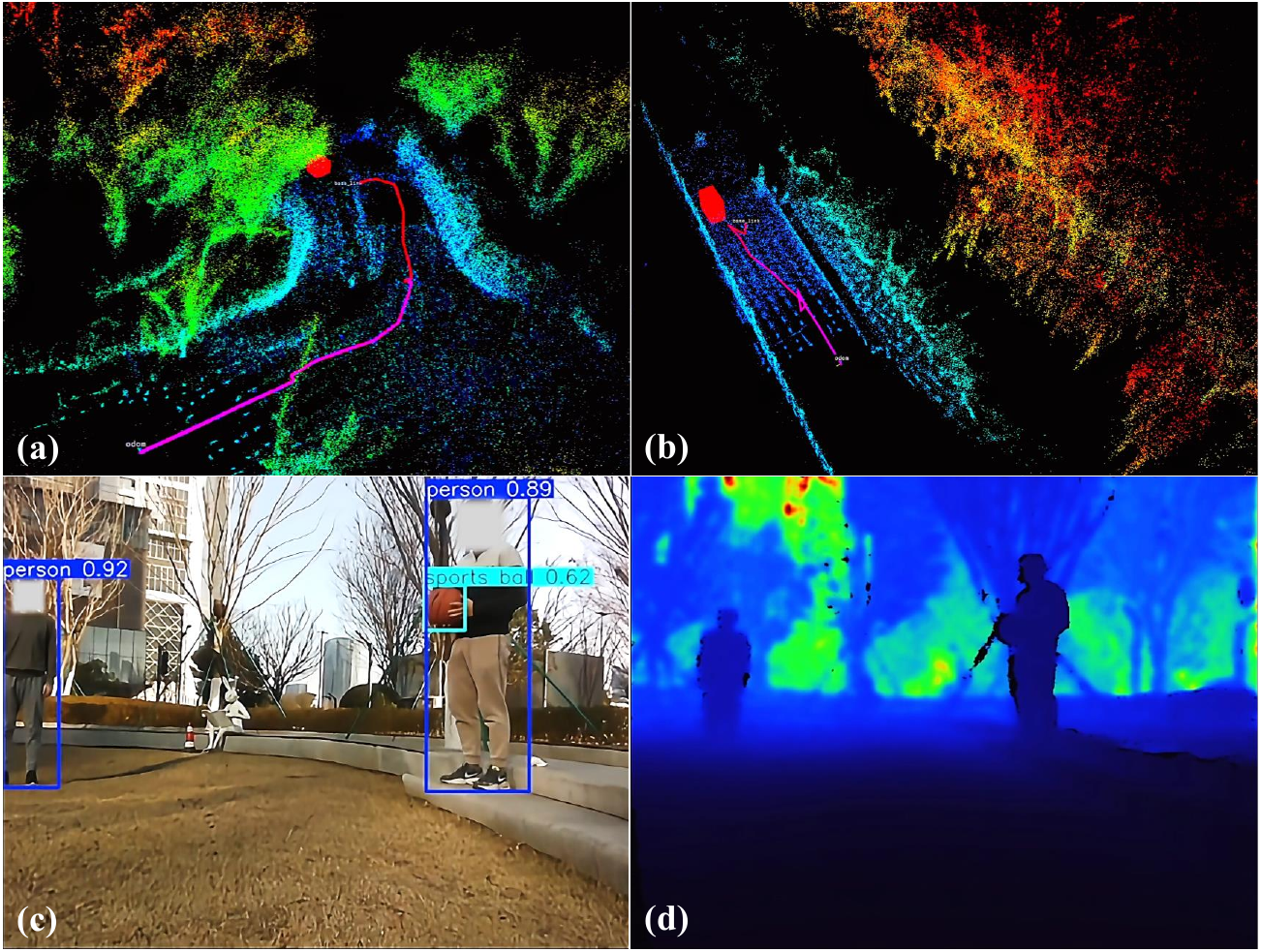}
    \caption{Onboard perception examples on the real robot. (a)(b) LiDAR point clouds from the omnidirectional 3D LiDAR scans. (c) RGB image. (d) Depth image.}
    \label{Fig: perception}
\vspace{-5mm}
\end{figure}

The real-robot system adopts the onboard perception pipeline from APREBot~\cite{xu2025aprebot} and runs it fully on Jetson Orin NX. We use a 360$^\circ$ LiDAR together with a forward-facing RGB-D camera in a complementary way: the LiDAR stream provides all-around coverage and continuous tracking, while the RGB-D stream is used to refine geometric estimates when the obstacle is observable in the frontal view. For each dynamic obstacle, the perception module outputs a metrically consistent state estimate including 3D position, velocity, and an effective size for downstream modules. Fig.~\ref{Fig: perception} shows example onboard sensing outputs from LiDAR point clouds, RGB images, and depth images.

\noindent\textbf{LiDAR-based omnidirectional sensing.}
Livox MID-360 is used to monitor moving objects around the robot. Each scan is aligned to a common world frame using LiDAR--inertial localization and then simplified by basic filtering, including region-of-interest selection and downsampling. We identify obstacle point groups by DBSCAN clustering and keep their identities consistent over time through Hungarian matching with track initialization and termination. A lightweight 3D Kalman filter is applied to suppress jitter and to produce stable obstacle position, velocity, and effective size estimates.

\noindent\textbf{RGB-D-based frontal refinement.}
Intel RealSense D435 provides refined estimates when an obstacle falls inside the camera field of view. We run YOLO on RGB images to obtain detections and use ByteTrack to connect detections across time. For each tracked object, we extract a compact depth region by growing from the track center on the depth image, and then lift the selected depth pixels into a local 3D point set using calibrated camera parameters. The refined 3D centroid and an effective radius are computed from this point set, and the obstacle velocity is obtained from consecutive refined 3D positions, complementing the LiDAR-based tracking.

\subsection{Simulation randomization settings}

We apply domain randomization during simulation training to improve robustness to modeling mismatch and sensing imperfections. Tab.~\ref{tab:domain_rand} summarizes the randomized factors and their ranges.

The randomization includes bounded perturbations on robot state estimates and obstacle observations, episode-level variations on obstacle initial conditions and speed, as well as system-level variations on ground contact and end-to-end latency.

\begin{table}[t]
    \centering
    \caption{Domain randomization used in simulation training.}
    \label{tab:domain_rand}
    \setlength{\tabcolsep}{7pt}
    \begin{tabular}{l l}
        \toprule
        Randomized factor & Value \\
        \midrule
        Base position noise (per axis) & $\mathcal{U}(-0.03,0.03)$~m \\
        Base yaw noise & $\mathcal{U}(-2,2)$~deg \\
        Base linear velocity noise (per axis) & $\mathcal{U}(-0.10,0.10)$~m/s \\
        Base angular velocity noise (per axis) & $\mathcal{U}(-0.15,0.15)$~rad/s \\
        Obstacle initial position (per axis) & $\mathcal{U}(-3.0,3.0)$~m \\
        Obstacle speed & $\mathcal{U}(0.5,4.0)$~m/s \\
        Obstacle position noise (per axis) & $\mathcal{U}(-0.05,0.05)$~m \\
        Obstacle velocity noise (per axis) & $\mathcal{U}(-0.20,0.20)$~m/s \\
        Observation dropout & $p=0.05$, hold $1$--$3$ steps \\
        Ground friction factor & $\mathcal{U}(0.6,1.2)$ \\
        End-to-end latency & $\mathcal{U}(0.00,0.06)$~s \\
        \bottomrule
    \end{tabular}
\vspace{-5mm}
\end{table}

\subsection{Additional experimental illustrations}~\label{Subsec: Additional experimental illustrations}
\vspace{-4mm}

We present additional qualitative results from simulation and real-robot experiments to showcase a wider range of evaluation scenarios.

\begin{figure*}[t]
    \centering
    \includegraphics[width=0.97\linewidth]{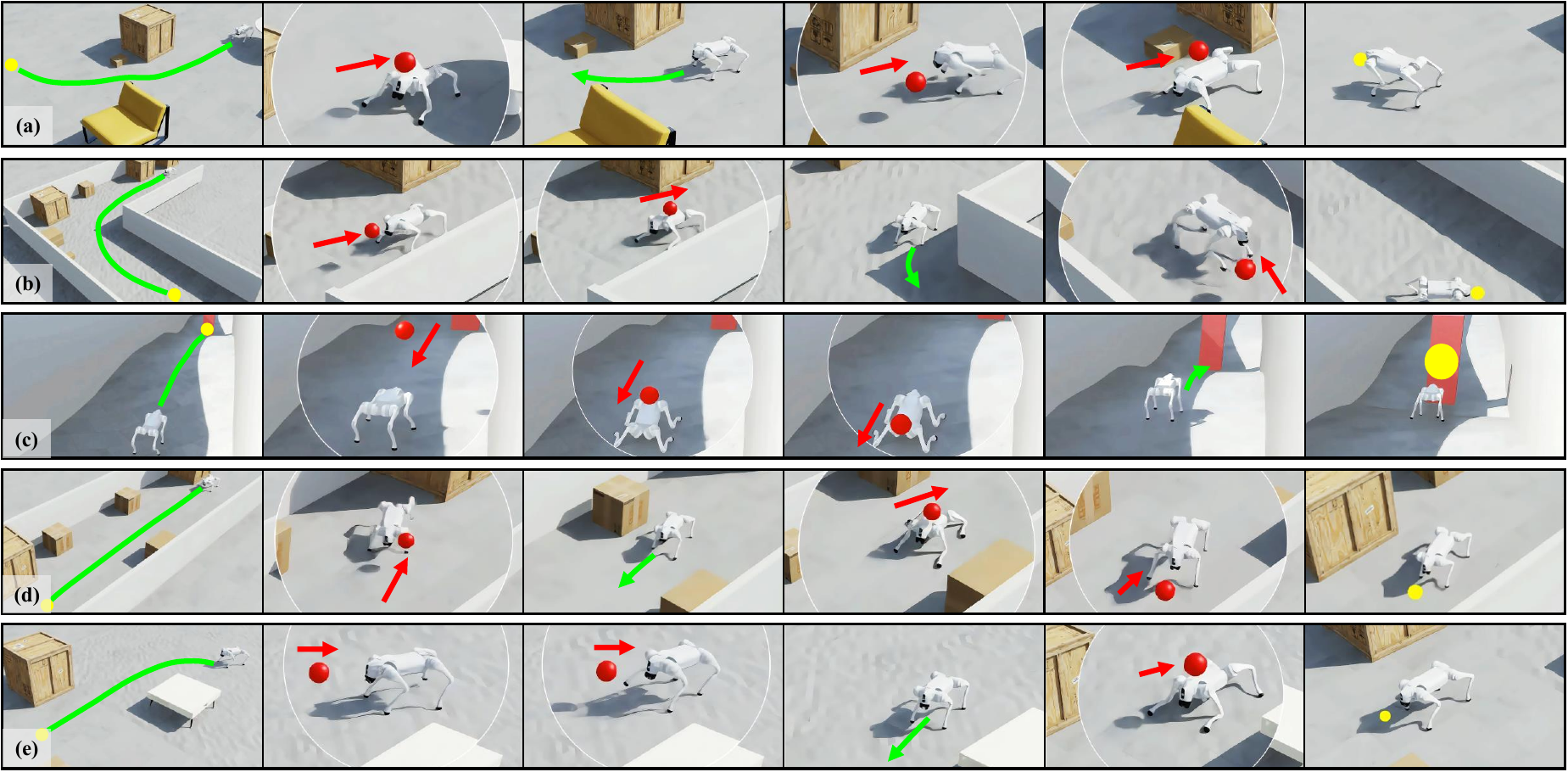}
    \caption{\textbf{Additional simulation scenarios.} Extended qualitative results in simulation across diverse terrain and space constraints. (a) Open flat ground with static obstacles. (b) Confined corridor with rough terrain. (c) Slope terrain. (d) Confined corridor on flat ground. (e) Open space combining flat and rough terrain.}
    \label{Fig: App exp sim}
\vspace{-5mm}
\end{figure*}

In simulation, we include extended scenarios that vary both terrain and spatial constraints, covering open rooms and confined corridors with flat ground, rough terrain, and slopes (Fig.~\ref{Fig: App exp sim}).

\begin{figure*}[th]
    \centering
    \includegraphics[width=0.97\linewidth]{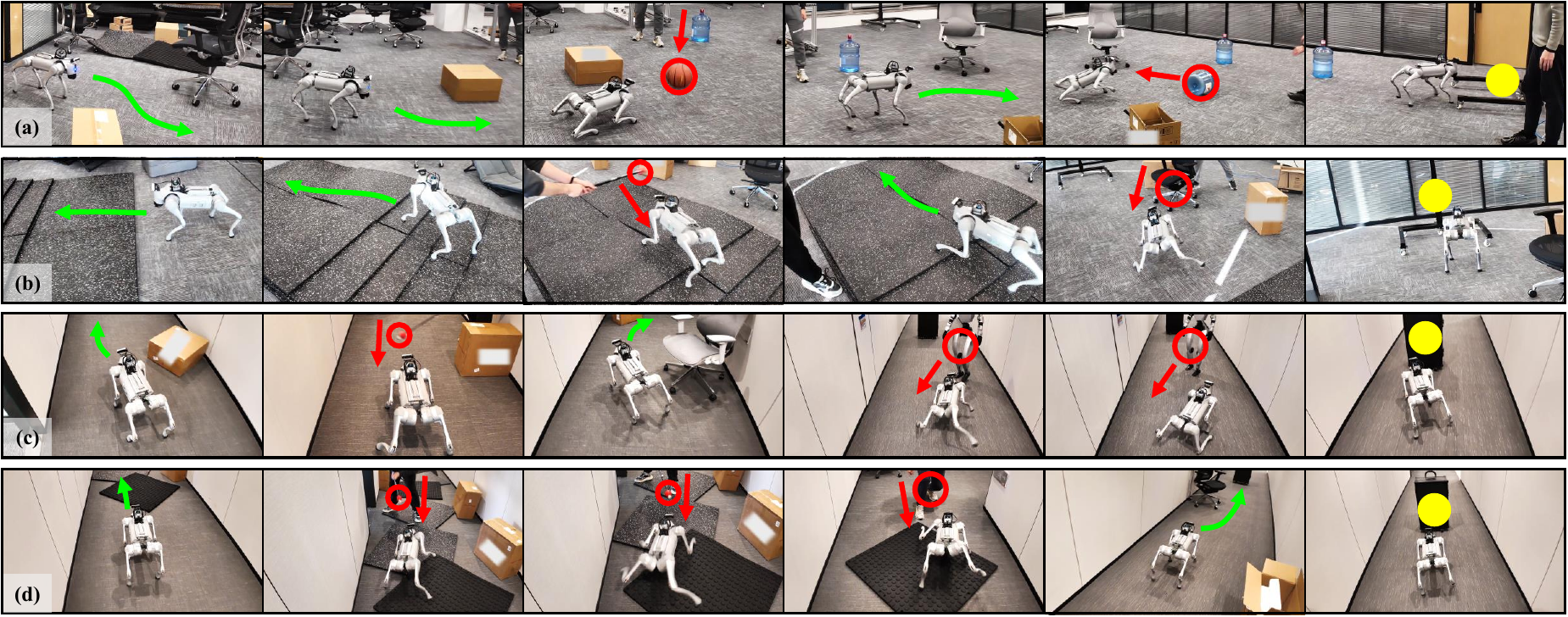}
    \caption{\textbf{Additional indoor real-robot experiments.} Extended indoor trials across open rooms and corridors with different terrain conditions. (a) Open indoor room on flat ground. (b) Open indoor room with step terrain. (c) Indoor corridor on flat ground. (d) Indoor corridor with rough terrain.}
    \label{Fig: App exp real indoor}
\vspace{-5mm}
\end{figure*}

On the real robot, we further present additional indoor trials in open rooms and corridors under different terrain conditions, including flat ground, step terrain, and rough surfaces (Fig.~\ref{Fig: App exp real indoor}).

\begin{figure*}[th]
    \centering
    \includegraphics[width=0.97\linewidth]{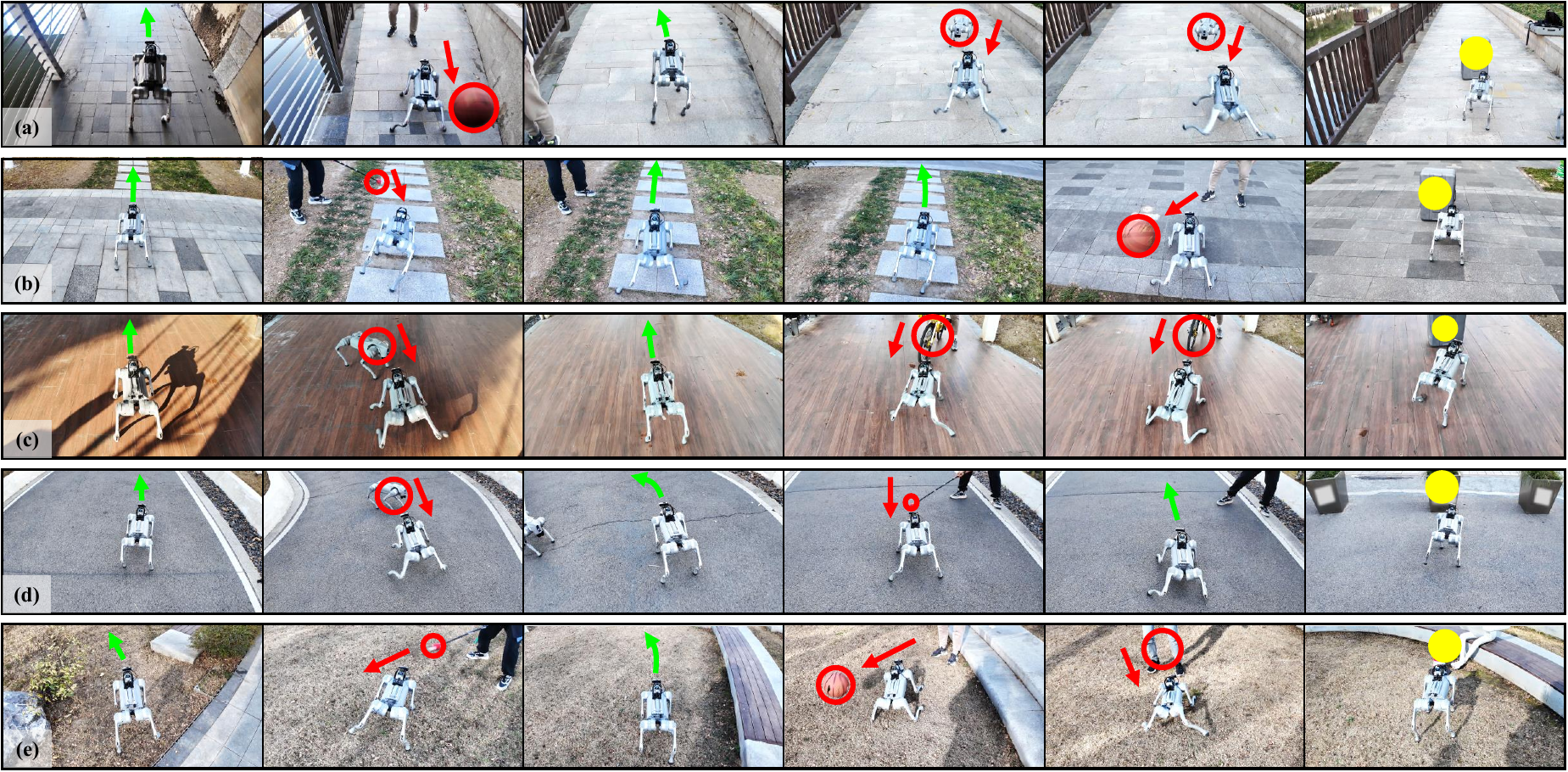}
    \caption{\textbf{Additional outdoor real-robot experiments.} Extended outdoor trials across diverse terrains and ground profiles. (a) Riverside trail. (b) Stone-paved path. (c) Wooden floor. (d) Slope terrain. (e) Grass field.}
    \label{Fig: App exp real outdoor}
\vspace{-5mm}
\end{figure*}

We also include extended outdoor trials across diverse terrains, including riverside trails, stone-paved paths, wooden floors, slopes, and grass fields (Fig.~\ref{Fig: App exp real outdoor}).

